\documentclass[11pt]{article}

\usepackage{acl}
\usepackage{times}
\usepackage{latexsym}
\usepackage{caption}
\usepackage{amsmath, amssymb}
\usepackage{algorithm}
\usepackage{algpseudocode}
\usepackage{booktabs}
\usepackage{enumitem}
\usepackage{xcolor}
\usepackage{subcaption}
\usepackage{multirow}
\usepackage{pgfplots}
\pgfplotsset{compat=1.18}

\usepackage[T1]{fontenc}

\usepackage[utf8]{inputenc}

\usepackage{microtype}

\usepackage{inconsolata}

\usepackage{graphicx}
\usepackage[most]{tcolorbox}
\usepackage{xcolor}
\usepackage{enumitem}

\newtcolorbox{promptbox}[2][]{
    enhanced,
    breakable,
    colback=gray!3,
    colframe=gray!55,
    boxrule=0.5pt,
    arc=1.5mm,
    left=6pt,
    right=6pt,
    top=6pt,
    bottom=6pt,
    before skip=6pt,
    after skip=8pt,
    fonttitle=\bfseries,
    coltitle=black,
    title={#2},
    #1
}
\newcommand{\placeholder}[1]{\textcolor{blue!55!black}{\texttt{\{#1\}}}}
\newcommand{\scoreset}{\texttt{\{0, 0.25, 0.5, 0.75, 1.0\}}}

\title{Learning What to Learn: Stage-Specific Data Sets for SFT-then-RL in Small Language Model Reasoning}

\author{
  Chongyang He$^{1}$\thanks{Equal contribution.}\thanks{Corresponding author.}
  \quad Rui Zhang$^{2}$\footnotemark[1]
  \quad Zixun Wang$^{3}$
  \quad Xin Li$^{4}$ \\
  $^{1}$Tsinghua University
  \quad $^{2}$National University of Singapore \\
  \quad $^{3}$DiDi
  \quad $^{4}$University of Electronic Science and Technology of China \\
  \texttt{hecy25@mails.tsinghua.edu.cn}
}

\begin{document}
\maketitle
\begin{abstract}
Post-training Small Language Models (SLMs) for reasoning typically follows an SFT-then-RL pipeline, yet existing work rarely considers what data should be learned at each stage. 
We argue that data strategy should be aligned with the distinct roles of SFT and RL: SFT is better suited for acquiring not-yet-mastered reasoning skills, while RL is better suited for consolidating skills that the model can already partially access. 
Based on this principle, we propose a difficulty-aware SFT-then-RL framework that organizes training data into stage-specific sets. 
For hard samples in the SFT stage, we introduce a Bridge mechanism that transforms raw teacher-generated reasoning traces into more learnable supervision for SLMs. 
For hard samples that remain unsolved during RL, we apply Critique Fine-Tuning by converting all-zero-reward failures into diagnostic, repair, and new reasoning trace supervision for the next SFT stage. 
Experiments on two SLMs across five reasoning benchmarks show that our method consistently improves over representative SFT, distillation, and RL baselines. 
Our results highlight the importance of coordinating data difficulty across SFT and RL for effective SLM reasoning post-training.
\end{abstract}

\section{Introduction}

The emergence of reasoning-oriented Large Language Models (LLMs), such as DeepSeek-R1~\cite{Guo_2025} and OpenAI o1~\cite{openai2026openaio1card}, has shown that Supervised Fine-Tuning followed by Reinforcement Learning (\textit{SFT-then-RL}) can substantially enhance multi-step reasoning capabilities. 
Small Language Models (SLMs; $\leq$ 3B parameters) offer complementary advantages in efficiency, cost, and accessibility, making them attractive for resource-constrained applications. 
While recent studies have explored reasoning distillation and post-training for SLMs~\cite{chen2025skipthinkingchunkwisechainofthoughtdistillation,guan2025recallextenddynamicsenhancingsmall,luo2025valleypatheffectivelong}, a fundamental question remains under-explored: \textit{how should reasoning data of different difficulty levels be allocated across the SFT and RL stages for SLMs?}

Prior work~\cite{guan2025recallextenddynamicsenhancingsmall} suggests that the two stages in SFT-then-RL play different functional roles. 
SFT can be viewed as an \textit{Extend} stage, where CoT distillation expands the model's reasoning boundary by introducing new reasoning patterns from stronger teacher models~\cite{Guo_2025,kim2025reinforcementlearningvsdistillation}. 
In contrast, RL with verifiable rewards is often better understood as a \textit{Recall} stage: rather than imparting entirely new capabilities, it primarily amplifies and optimizes knowledge already accessible to the model~\cite{shao2024deepseekmathpushinglimitsmathematical,yue2025doesreinforcementlearningreally}. 
This stage-specific view implies that SFT should expose SLMs to reasoning skills that are not yet fully mastered but remain learnable, whereas RL should focus on samples for which the model can already discover correct solutions with non-zero probability.

However, directly transferring hard reasoning data to SLMs is non-trivial. 
Recent studies show that SLMs often struggle to learn from long and complex teacher-generated CoT traces, a phenomenon known as \textit{Long CoT Degradation}~\cite{luo2025valleypatheffectivelong}. 
In such cases, overly complex CoT supervision may cause SLMs to learn superficial reasoning patterns rather than robust problem-solving skills~\cite{li2025smallmodelsstrugglelearn}. 
Moreover, LLM-generated reasoning chains often contain logical jumps and redundancy~\cite{cai2025enhancingreasoningabilitiessmall}, further increasing the mismatch between teacher supervision and the learning capacity of SLMs. 
These findings suggest that the key challenge is not simply whether hard samples should be used, but how they should be transformed and assigned to different post-training stages.

\begin{figure}[t]
  \includegraphics[width=\columnwidth]{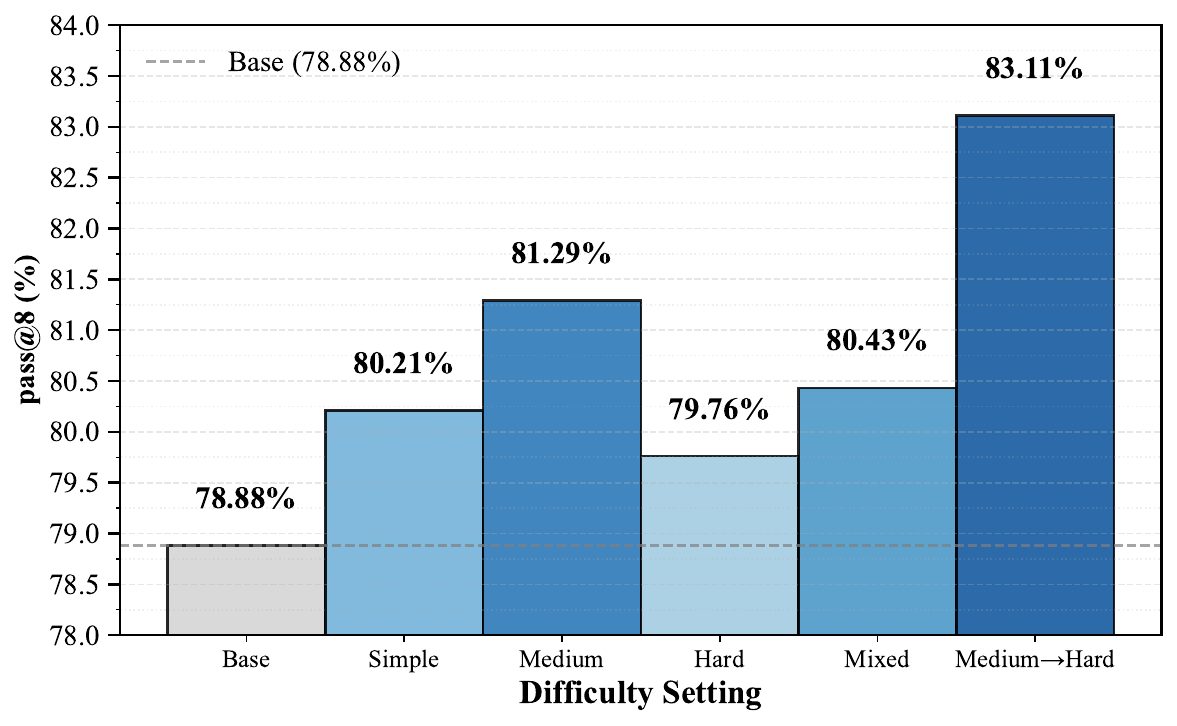}
  \caption{
  Motivating difficulty-aware SFT experiment on Llama3.2-1B-Instruct. 
  Under the same token budget, the Medium$\rightarrow$Hard curriculum achieves the strongest post-SFT pass@8 performance.
  }
  \label{fig:intro}
\end{figure}

Motivated by this question, we conduct controlled experiments on the GSM8K training split to examine how sample difficulty affects SFT-stage learning. 
Across both Qwen2.5-0.5B-Instruct and Llama3.2-1B-Instruct, we observe a consistent pattern: under the same supervised token budget, a Medium$\rightarrow$Hard curriculum yields stronger post-SFT reasoning performance than training on a single difficulty group or a mixed set. 
We adopt pass@8 as the primary metric for evaluating the SFT stage, as prior work~\cite{kang2025quagmiressftrlposttraininghigh} suggests that pass@$k$ more reliably predicts downstream performance following RL in the \textit{SFT-then-RL} paradigm. 
As shown in Figure~\ref{fig:intro}, this result suggests that moderately challenging samples provide a strong acquisition signal, while hard samples may further help only when introduced after a learnable starting point. 
Detailed settings of this motivating experiment are provided in Appendix~\ref{app:motivating_experiment}.

Importantly, this observation does not imply that hard samples should be discarded. 
Hard samples often contain reasoning patterns that go beyond the current capability of SLMs and are therefore valuable for expanding its reasoning boundary. 
The challenge is that raw hard samples distilled from stronger LLMs are often too long, too jumpy, or locally too difficult for SLMs to absorb directly. 
Thus, the problem lies not in hard samples themselves, but in the mismatch between raw LLM-generated reasoning trajectories and SLM learnability.

To address this challenge, we propose a difficulty-aware SFT-then-RL framework organized around three functional data sets. 
The \textit{Acquisition Set} is used for initial SFT and consists of medium samples and bridge-adjusted hard samples. 
To make hard samples learnable for SLMs, our \textit{Bridge} mechanism evaluates each reasoning step by \textit{importance}, \textit{jumpiness}, and \textit{difficulty}, and then applies keep, expand, compress, drop, or localize operations to transform raw hard reasoning traces into capacity-aligned supervision. 
After SFT, SLMs still encounter problems that they cannot solve, which leads to sparse rewards if directly used in RL. 
We therefore define the \textit{Consolidation Set} as all non-all-zero samples 
All-zero-reward samples are excluded from direct RL because the current policy fails to produce any correct solution under the rollout budget. 
Rather than handling these failures within RL through additional search or hinting~\cite{zhang2025breadbranchedrolloutsexpert}, we follow our stage-specific principle: skills inaccessible to the SLM should be learned through SFT. 
Thus, we build a \textit{Recycled Set}, where a stronger teacher converts all-zero failures into diagnostic, repair, and a new reasoning trajectory for the next SFT stage. 
This creates an iterative loop in which SFT expands the model's reasoning boundary and RL consolidates partially accessible skills.

Our contributions are summarized as follows:
\begin{itemize}
    \item We reveal that hard samples can provide stronger SFT gains when converted into capacity-aligned supervision and incorporated through a suitable curriculum, turning hard data into useful acquisition signals for SLMs.
    \item We propose a Bridge mechanism that converts hard reasoning traces into capacity-aligned supervision through step-level importance, jumpiness, and difficulty estimation.
    \item We introduce an error-guided recycling mechanism that transforms all-zero-reward RL failures into SFT supervision, and validate the full framework on two SLMs across in-domain and out-of-domain reasoning benchmarks.
\end{itemize}


\section{Related Work}

\subsection{Post-Training Paradigms for SLMs}

Recent reasoning-oriented LLMs, such as DeepSeek-R1~\citep{Guo_2025} and OpenAI o1~\citep{openai2026openaio1card}, have shown that \textit{SFT-then-RL} paradigm is effective for improving multi-step reasoning. 
For SLMs, this paradigm typically relies on distilling reasoning traces from stronger teacher models and then applying reward-based optimization with verifiable signals~\citep{zhang2025making,hu2026dereason}. 
Recent work further suggests that SFT and RL play different functional roles: SFT can extend the model's reasoning boundary by introducing new patterns, while RL mainly recalls and reinforces capabilities already accessible to the model~\citep{guan2025recallextenddynamicsenhancingsmall,shao2024deepseekmathpushinglimitsmathematical,yue2025doesreinforcementlearningreally}. 
Meanwhile, pass@large-$k$ has been shown to better predict downstream RL potential than greedy accuracy~\citep{kang2025quagmiressftrlposttraininghigh}. 
However, existing studies primarily focus on training algorithms~\cite{zhang2025breadbranchedrolloutsexpert,wang2025efficient}, leaving the stage-specific allocation of data difficulty underexplored.

\subsection{Capacity-Aligned Supervision for SLMs}

A major obstacle in reasoning distillation is the mismatch between complex LLM-generated reasoning traces and the limited learning capacity of SLMs. 
Directly learning from long and dense teacher traces can cause \textit{Long CoT Degradation}, where SLMs fail to internalize robust reasoning patterns~\citep{luo2025valleypatheffectivelong,li2025smallmodelsstrugglelearn}. 
To address this, prior work restructures or simplifies reasoning supervision through chunk-wise distillation~\citep{chen2025skipthinkingchunkwisechainofthoughtdistillation}, critique-and-refinement~\citep{cai2025enhancingreasoningabilitiessmall}, problem-space mapping~\citep{wang2025decouplingunderstandingreasoningproblem}, or selective reasoning~\citep{wang2025thinknotselectivereasoning}. 
For RL-stage failures, BREAD~\citep{zhang2025breadbranchedrolloutsexpert} introduces expert anchors to alleviate reward sparsity. 
While these methods adapt reasoning supervision or improve RL exploration, they primarily optimize either the SFT or RL stage in isolation. 
They still lack a coordinated data strategy that jointly considers what SLMs should acquire during SFT and what they should consolidate during RL.

\begin{figure*}[t]
    \centering
    \includegraphics[width=\textwidth]{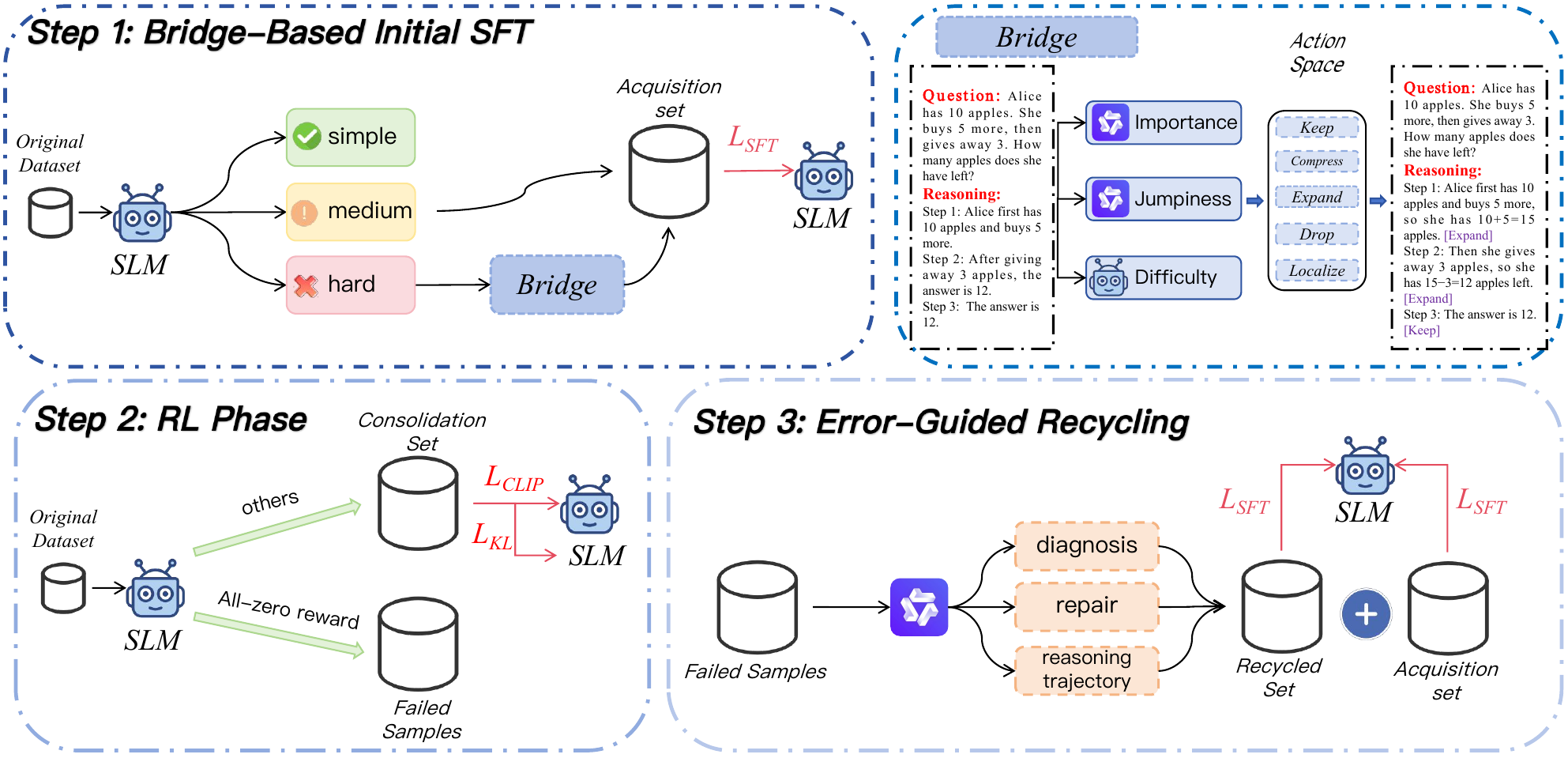}
    \caption{
    Overview of the proposed difficulty-aware SFT-then-RL framework. 
    The initial SFT stage builds an Acquisition Set from medium samples and bridge-adjusted hard samples. 
    The RL stage trains on the Consolidation Set, i.e., non-all-zero samples that provide usable reward signals for GRPO. 
    All-zero-reward failures are transformed into a Recycled Set through teacher-guided diagnosis, repair, and new reasoning trajectory supervision, and are fed back into the next SFT stage for iterative acquisition--consolidation.
    }
    \label{fig:workflow}
\end{figure*}

\section{Methodology}
\label{sec:methodology}

We develop a difficulty-aware post-training pipeline for SLM mathematical reasoning. We begin by partitioning training samples by model-specific difficulty and construct an initial SFT curriculum from medium samples and bridge-adjusted hard samples. 
We then perform RL on samples with informative reward signals, excluding all-zero-reward failures from direct policy optimization. 
These failures are recycled into the next SFT stage through teacher-guided diagnosis, repair, and a new reasoning trajectory supervision, forming an iterative loop where SFT expands missing skills and RL consolidates partially accessible ones.

\subsection{Stage-Specific Data Partitioning}
\label{sec:data_partition}

Given a training dataset $\mathcal{D}=\{(q_j,a_j^*)\}_{j=1}^{N}$, where $q_j$ denotes the $j$-th math problem and $a_j^*$ denotes its ground-truth answer, we estimate sample difficulty with respect to the current SLM $\pi_{\theta}$. 
For each question $q_j$, we sample $n$ responses:
\begin{equation}
    \{o_{j,i}\}_{i=1}^{n} \sim \pi_{\theta}(\cdot \mid q_j).
\end{equation}
Each response is evaluated by a rule-based verifier:
\begin{equation}
    r_{j,i} = \mathbb{I}[\mathrm{Ans}(o_{j,i})=a_j^*],
\end{equation}
where $\mathrm{Ans}(\cdot)$ extracts the final numerical answer from the response. 
We then compute the empirical correctness rate:
\begin{equation}
    \rho(q_j) = \frac{1}{n}\sum_{i=1}^{n} r_{j,i}.
\end{equation}
In our experiments, we set $n=64$ for difficulty estimation.

Based on $\rho(q_j)$, we partition the training samples into three model-specific difficulty groups: 
$\mathcal{D}_{simple}$ contains samples with $\rho(q_j)\in[0.75,1]$, 
$\mathcal{D}_{medium}$ contains samples with $\rho(q_j)\in[0.25,0.75)$, and 
$\mathcal{D}_{hard}$ contains samples with $\rho(q_j)\in[0,0.25)$.

Our motivating experiments show that a medium-to-hard SFT curriculum provides a stronger initialization for subsequent post-training. 
However, the raw form of hard samples is often unsuitable for SFT, as teacher-distilled reasoning traces can be long, abstract, and difficult for SLMs to absorb directly. 
We therefore apply a bridge-based transformation to convert hard samples into more learnable supervision, as detailed in Section~\ref{sec:bridge_sft}.

Accordingly, we organize the data into three functional sets. 
The \textbf{Acquisition Set} is used in initial SFT and consists of raw medium samples and bridge-adjusted hard samples. 
The \textbf{Consolidation Set} is used in RL and consists of non-all-zero samples, including mixed-reward samples that provide explicit reward contrast and all-one-reward samples that stabilize already accessible skills. 
Motivated by Critique Fine-Tuning (CFT)~\citep{wang2501critique,wang2025unleashing}, the \textbf{Recycled Set} consists of all-zero-reward samples that are unsuitable for direct RL and are instead converted into error-guided SFT supervision. 
This partition follows a stage-specific learning principle: SFT is better suited for acquiring missing reasoning skills, whereas RL is better suited for consolidating skills that the model can already partially access.

\subsection{Bridge-Based Initial SFT}
\label{sec:bridge_sft}

The goal of bridge-based initial SFT is to make hard samples learnable for SLMs without forcing them to directly imitate raw LLM-distilled reasoning traces. 
We adopt a \textit{medium-first, then bridged-hard} strategy: the model is first trained on $\mathcal{D}_{medium}$ to acquire manageable reasoning patterns, and then trained on a bridge-adjusted version of $\mathcal{D}_{hard}$, denoted as $\mathcal{D}_{bridge}$. 
The bridge transformation restructures hard reasoning traces by preserving critical steps, expanding abrupt transitions, compressing redundant content, dropping low-value difficult steps, and localizing steps that are difficult for the SLM to learn directly. Figure~\ref{fig:workflow} illustrates the overall workflow of our proposed framework.

\paragraph{Step-level decomposition.}
For each hard sample $(q,a^*) \in \mathcal{D}_{hard}$, we first obtain a teacher-generated reasoning trace $C$ from a stronger LLM. 
We then decompose $C$ into a sequence of reasoning steps:
\begin{equation}
    C = [s_1, s_2, \ldots, s_m],
\end{equation}
where each $s_i$ corresponds to a coherent reasoning unit in the solution process. 
Each step is subsequently evaluated along three dimensions: importance, jumpiness, and difficulty.

\paragraph{Importance.}
The importance score measures how necessary a reasoning step is for supporting the final answer. 
For each step $s_i$, we remove it from the original reasoning trace and ask a judge model to evaluate how much the remaining trace is damaged:
\begin{equation}
    I_i = J_{\mathrm{imp}}(q, a^*, C, C_{\setminus i}, s_i),
\end{equation}
where $C_{\setminus i}$ denotes the reasoning trace after deleting $s_i$, and $J_{\mathrm{imp}}$ is a teacher judge. 
The score $I_i \in \{0,0.25,0.5,0.75,1.0\}$ reflects whether the step is redundant, moderately useful, or critical to the reasoning process. 
A higher score indicates that deleting the step substantially weakens the remaining trace's ability to justify the correct answer.

\paragraph{Jumpiness.}
The jumpiness score measures whether the transition from previous reasoning steps to the current step is too abrupt for an SLM to follow. 
For step $s_i$, we define its previous context as:
\begin{equation}
    P_i = [s_1,\ldots,s_{i-1}].
\end{equation}
A teacher judge evaluates the logical gap between $P_i$ and $s_i$:
\begin{equation}
    J_i = J_{\mathrm{jump}}(q, P_i, s_i),
\end{equation}
where $J_i \in \{0,0.25,0.5,0.75,1.0\}$. Since the first step has no preceding context, we set $J_1=0$. A higher jumpiness score indicates a larger logical gap between the preceding context and the current step, suggesting that additional intermediate reasoning is needed for an SLM to follow the transition. The prompt template used for jumpiness scoring is provided in Appendix~\ref{bridge_trace}.

\paragraph{Difficulty.}
The difficulty score measures how hard a step is for the current SLM to imitate under teacher forcing. 
Given the question $q$ and the previous steps $P_i$, we compute the average token-level negative log-likelihood of the current step under the SLM:
\begin{equation}
    D_i = - \frac{1}{|s_i|} \sum_{t=1}^{|s_i|}
    \log \pi_{\theta}(s_{i,t} \mid q, P_i, s_{i,<t}).
\end{equation}
This score captures the local learnability of each step. 
A higher $D_i$ indicates that the current SLM assigns lower probability to the step, suggesting that the step may be too difficult to learn directly. 

\paragraph{Bridge operations.}
Based on the three scores $(I_i,J_i,D_i)$, we apply a step-level transformation to each hard reasoning trace. 
The transformation is selected from five operations. 
\textbf{(1) Keep} preserves an important, smooth, and locally learnable step unchanged. 
\textbf{(2) Compress} shortens a less important but still useful step, or a step with redundant explanations, while preserving its core meaning. 
\textbf{(3) Expand} adds missing intermediate reasoning for an important but jumpy step to smooth the transition from previous steps. 
\textbf{(4) Drop} removes an unimportant step that is either too jumpy or too difficult, reducing noisy or low-value supervision. 
\textbf{(5) Localize} rewrites an important but locally difficult step into an easier-to-learn form and constructs an additional local SFT sample that trains the model to predict this step from its preceding context.

Formally, the bridge transformation can be written as:
\begin{equation}
    \tilde{s}_i = \mathcal{B}(s_i; I_i, J_i, D_i, q, P_i),
\end{equation}
where $\mathcal{B}$ denotes the bridge operation applied to step $s_i$. 
The full decision procedure is provided in Appendix~\ref{bridged_samples}. 
After processing all steps, we obtain a bridge-adjusted reasoning trace:
\begin{equation}
    \tilde{C} = [\tilde{s}_1,\tilde{s}_2,\ldots,\tilde{s}_{m'}],
\end{equation}
where $m'$ may differ from $m$ when steps are dropped or locally restructured. 
The resulting bridge-adjusted hard dataset is denoted as:
\begin{equation}
    \mathcal{D}_{bridge} = \{(q,\tilde{C},a^*) \mid (q,a^*) \in \mathcal{D}_{hard}\}.
\end{equation}

\paragraph{Initial SFT curriculum.}
The initial SFT stage follows a two-step curriculum. 
We first train the model on $\mathcal{D}_{medium}$ and then continue training on $\mathcal{D}_{bridge}$. 
Equivalently, the initial Acquisition Set is:
\begin{equation}
    \mathcal{D}_{acq}
    =
    \mathcal{D}_{medium}
    \cup
    \mathcal{D}_{bridge},
\end{equation}
with the training order:
\begin{equation}
    \mathcal{D}_{medium}
    \rightarrow
    \mathcal{D}_{bridge}.
\end{equation}
In this way, $\mathcal{D}_{medium}$ provides stable and learnable supervision, while $\mathcal{D}_{bridge}$ introduces harder reasoning patterns in a form better aligned with the learning capacity of SLMs.

\subsection{Error-Guided Recycling for Iterative SFT}
\label{sec:error_guided_recycling}

After the initial SFT stage, the model enters RL for reward-guided consolidation. 
However, not all samples are suitable for direct RL. 
If the current policy cannot produce any correct solution for a sample under the rollout budget, the resulting all-zero rewards provide little useful signal for GRPO. 
To handle such RL-intractable failures, we introduce an \textit{error-guided recycling} mechanism that converts them into SFT-learnable supervision.

\paragraph{Reward-based sample partitioning.}
In the RL stage, we adopt GRPO~\cite{shao2024deepseekmathpushinglimitsmathematical} for policy optimization. 
For each candidate question $q \in \mathcal{D}_{cand}^{(t)}$, the current policy generates $G$ rollouts, each assigned a rule-based reward. 
Let $\mathbf{r}(q)=[r_1,\ldots,r_G]$ denote the reward vector. 
We partition the candidate samples into the Consolidation Set and the failed set:
\begin{equation}
\begin{aligned}
\mathcal{D}_{con}^{(t)}
&=
\left\{(q,a^*) \mid \sum_{i=1}^{G} r_i > 0 \right\}, \\
\mathcal{D}_{fail}^{(t)}
&=
\left\{(q,a^*) \mid \sum_{i=1}^{G} r_i = 0 \right\}.
\end{aligned}
\end{equation}

The Consolidation Set contains all non-all-zero samples, including mixed-reward samples that provide explicit reward contrast and all-one-reward samples that may still produce zero-reward rollouts during RL training. 
In contrast, samples in $\mathcal{D}_{fail}^{(t)}$ receive uniformly zero rewards during filtering, and we empirically find that they rarely produce positive rewards in subsequent RL training. 
They therefore provide sparse and uninformative feedback for GRPO. 
Rather than directly optimizing on these samples, we exclude them from RL and recycle them into the next SFT stage.

\paragraph{Representative negative response selection.}
For each failed sample $(q,a^*) \in \mathcal{D}_{fail}^{(t)}$, all sampled responses are incorrect:
\begin{equation}
    \mathcal{O}^{-}(q)=\{o_i \mid r_i=0\}_{i=1}^{G}.
\end{equation}
Instead of analyzing every incorrect response, we select one representative negative response $o^{-}$ for teacher-guided diagnosis. 
We prefer \textit{near-miss} responses, where the model produces a partially meaningful reasoning process but deviates at a specific step. 
Such responses expose concrete reasoning gaps and are more useful than short, irrelevant, or degenerate outputs.

To reduce the cost of teacher-guided diagnosis, we do not analyze every incorrect rollout. 
Instead, we use a lightweight heuristic scoring function $S_{\mathrm{neg}}(\cdot)$, detailed in Appendix~\ref{app:negative_selection}, to select a near-miss response with sufficient reasoning content, explicit step structure, and a parseable final-answer format. 
For each failed sample, this yields a representative negative response $o^{-}$ for subsequent teacher-guided diagnosis.

\paragraph{Teacher-guided error analysis.}
Given the question $q$, ground-truth answer $a^*$, and representative negative response $o^{-}$, we query a stronger teacher model $T$ to generate a structured diagnosis:
\begin{equation}
    e = T(q,a^*,o^{-}).
\end{equation}
The diagnosis focuses on the earliest reasoning failure rather than generating a long alternative solution. 
We represent it as:
\begin{equation}
    e =
    \big(
    f_{\mathrm{err}},
    c_{\mathrm{err}},
    w_{\mathrm{err}},
    k_{\mathrm{miss}},
    h_{\mathrm{fix}},
    s_{\mathrm{next}},
    c_{\mathrm{new}}
    \big),
\end{equation}
where $f_{\mathrm{err}}$ denotes the first erroneous step in $o^{-}$, $c_{\mathrm{err}}$ categorizes the error type, and $w_{\mathrm{err}}$ explains why the step is incorrect. 
The term $k_{\mathrm{miss}}$ describes the missing knowledge or reasoning pattern, $h_{\mathrm{fix}}$ provides a minimal correction hint, $s_{\mathrm{next}}$ gives the correct next step after the error point, and $c_{\mathrm{new}}$ denotes a concise corrected reasoning trajectory. 
This structured diagnosis converts each failed RL sample into targeted supervision for next SFT stage.

\paragraph{Recycled SFT sample construction.}
Given the structured diagnosis $e$, we construct three types of recycled SFT supervision: diagnostic supervision for identifying the first error, repair supervision for correcting the local reasoning gap, and new-reasoning trace supervision for recovering a concise solution.

Let $p_{\mathrm{pre}}$ denote the reasoning prefix before the first error in $o^{-}$. 
We define:
\begin{equation}
\begin{aligned}
\mathcal{D}_{diag}^{(t)} &: (q,o^{-}) \mapsto (c_{\mathrm{err}}, f_{\mathrm{err}}, w_{\mathrm{err}}), \\
\mathcal{D}_{repair}^{(t)} &: (q,p_{\mathrm{pre}},f_{\mathrm{err}},h_{\mathrm{fix}}) \mapsto s_{\mathrm{next}}, \\
\mathcal{D}_{new}^{(t)} &: q \mapsto c_{\mathrm{new}}.
\end{aligned}
\end{equation}

For $\mathcal{D}_{new}^{(t)}$, we require $c_{\mathrm{new}}$ to end with the same final-answer format used by the reward verifier, i.e., \texttt{\#\#\#\# <answer>}. 
This ensures consistency between recycled SFT and subsequent RL rollouts. 
The resulting recycled dataset is:
\begin{equation}
    \mathcal{D}_{rec}^{(t)}
    =
    \mathcal{D}_{diag}^{(t)}
    \cup
    \mathcal{D}_{repair}^{(t)}
    \cup
    \mathcal{D}_{new}^{(t)}.
\end{equation}

\paragraph{Iterative recycling into SFT.}
At iteration $t+1$, the recycled dataset is incorporated into the next SFT stage together with the acquisition data:
\begin{equation}
    \mathcal{D}_{SFT}^{(t+1)}
    =
    \mathcal{D}_{acq}
    \cup
    \mathcal{D}_{rec}^{(t)}.
\end{equation}
Here, $\mathcal{D}_{acq}$ provides acquisition-oriented supervision, while $\mathcal{D}_{rec}^{(t)}$ injects targeted supervision derived from previous RL failures.

Overall, error-guided recycling converts all-zero-reward samples, which provide little useful signal for direct GRPO optimization, into structured supervision for the next SFT stage. 
This forms an iterative SFT-then-RL loop: RL exposes capability gaps through failed rollouts, while SFT repairs these gaps through diagnostic, repair, and reasoning trace supervision. 
In this way, SFT progressively expands the model's reasoning boundary, while RL consolidates reasoning skills that the model can already partially access.

\section{Experiments}
\subsection{Experimental Setup}

\paragraph{Datasets.}
Following~\citet{wang2025decouplingunderstandingreasoningproblem}, we train all models exclusively on the GSM8K training split~\cite{cobbe2021training}. 
For evaluation, we use GSM8K-Platinum~\cite{vendrow2025large} as the in-domain benchmark, and MAWPS~\cite{koncel2016mawps}, SVAMP~\cite{patel2021nlp}, MATH500~\cite{hendrycks2021measuring}, and LogiQA~\cite{liu2020logiqa} as out-of-domain benchmarks covering mathematical and logical reasoning.

\paragraph{Baselines.}
We compare our method with representative baselines covering CoT distillation, prompt optimization, RL-based training, and knowledge distillation. 
Specifically, we include Std-CoT~\cite{magister2023teaching} and STaR~\cite{zelikman2024star} for CoT distillation, PRewrite~\cite{kong2024prewrite} for prompt optimization, GRPO~\cite{shao2024deepseekmathpushinglimitsmathematical} as a direct RL baseline, and Vanilla-KD~\cite{muralidharan2024compact} for teacher-based knowledge distillation. 
Following prior work~\cite{sheng2025learning,wang2025decouplingunderstandingreasoningproblem}, we use answer accuracy as the primary evaluation metric.

\paragraph{Implementation Details.}
To evaluate the generalizability of our method, we conduct experiments on two representative SLMs: Qwen2.5-0.5B-Instruct~\cite{yang2025qwen3} and Llama3.2-1B-Instruct~\cite{grattafiori2024llama}. 
All experiments are conducted on 4 NVIDIA RTX 4090 GPUs with 24GB memory each. 
We implement post-training with VERL~\cite{sheng2025hybridflow}. 
During inference, we use greedy decoding and accelerate generation with vLLM~\cite{kwon2023efficient}. 
The prompt templates are provided in Appendix~\ref{app:prompt_templates}. 
Detailed hyperparameters and dataset statistics are reported in Appendices~\ref{app:training_hyperparameters} and~\ref{app:dataset_statistics}, respectively.

\begin{table*}[t]
\centering
\small
\setlength{\tabcolsep}{4.5pt}
\renewcommand{\arraystretch}{1.2}
\begin{tabular}{l c cccc c}
\toprule
\multirow{2}{*}{\textbf{Methods}} 
& \textbf{In-Domain} 
& \multicolumn{4}{c}{\textbf{Out-of-Domain}} 
& \multirow{2}{*}{\textbf{Average}} \\
\cmidrule(lr){2-2} \cmidrule(lr){3-6}
& GSM8K-Platinum & MAWPS & SVAMP & MATH500 & LogiQA & \\
\midrule
\multicolumn{7}{l}{\textit{\textbf{\# Qwen2.5-0.5B-Instruct based}}} \\
\midrule
Base~\cite{yang2025qwen3} & 45.74 & 54.23 & 54.67 & 27.80 & 14.44 & 39.38 \\
CoT-Dis~\cite{magister2023teaching} & 44.67 & 55.77 & 58.33 & 18.80 & \underline{30.41} & 41.60 \\
STaR~\cite{zelikman2024star} & 51.86 & 57.88 & 61.67 & 29.60 & 23.50 & 44.90 \\
GRPO~\cite{shao2024deepseekmathpushinglimitsmathematical} & 51.03 & 58.08 & 61.00 & 27.40 & 22.73 & 44.05 \\
PRewrite~\cite{kong2024prewrite} & 47.23 & 56.73 & 57.00 & 29.80 & 23.96 & 42.94 \\
Vanilla-KD~\cite{muralidharan2024compact} & 49.30 & 57.69 & 61.67 & 30.40 & 20.74 & 43.96 \\
DURIT~\cite{wang2025decouplingunderstandingreasoningproblem} & \underline{53.10} & \underline{60.38} & \underline{63.00} & \underline{32.80} & 25.81 & \underline{47.02} \\
Our Model & \textbf{54.51} & \textbf{63.94} & \textbf{64.00} & \textbf{33.60} & \textbf{32.41} & \textbf{49.69} \\
\midrule
\multicolumn{7}{l}{\textit{\textbf{\# Llama3.2-1B-Instruct based}}} \\
\midrule
Base~\cite{grattafiori2024llama} & 30.52 & 5.77 & 20.67 & 22.60 & 1.54 & 16.22 \\
CoT-Dis~\cite{magister2023teaching} & 48.06 & 56.92 & 57.67 & 24.60 & \underline{21.81} & 41.81 \\
STaR~\cite{zelikman2024star} & 36.31 & 52.50 & 54.33 & 20.00 & 8.45 & 34.32 \\
GRPO~\cite{shao2024deepseekmathpushinglimitsmathematical} & 48.39 & 59.23 & 57.67 & 26.40 & 4.45 & 39.23 \\
PRewrite~\cite{kong2024prewrite} & 35.81 & 41.34 & 46.00 & 18.80 & 3.53 & 29.10 \\
Vanilla-KD~\cite{muralidharan2024compact} & 42.35 & \underline{64.23} & 62.67 & 22.40 & 7.99 & 39.93 \\
DURIT~\cite{wang2025decouplingunderstandingreasoningproblem} & \underline{52.36} & 62.31 & \textbf{66.00} & \textbf{27.60} & 19.82 & \underline{45.62} \\
Our Model & \textbf{54.18} & \textbf{72.68} & \underline{65.67} & \underline{26.60} & \textbf{23.50} & \textbf{48.53} \\
\bottomrule
\end{tabular}
\caption{Performance (\%) of Qwen2.5-0.5B-Instruct and Llama3.2-1B-Instruct models across five representative benchmarks under various methods. The \textbf{bold} and \underline{underline} indicate the best and second-best results within each model family, respectively.}
\label{table:main_results}
\end{table*}

\subsection{Experimental Results}

\begin{figure}[t]
  \centering
  \includegraphics[width=\columnwidth]{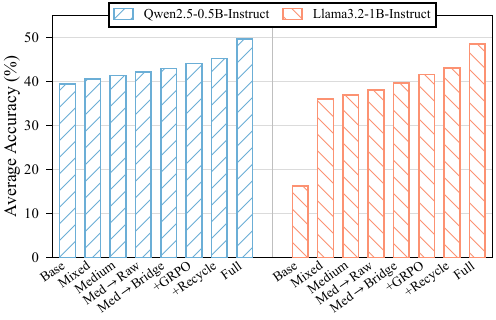}
  \caption{
  Component ablation results averaged across five evaluation benchmarks. 
  The results show the contribution of each stage-specific data strategy.
  }
  \label{fig:component_ablation}
\end{figure}

\begin{figure}[t]
  \centering
  \includegraphics[width=\columnwidth]{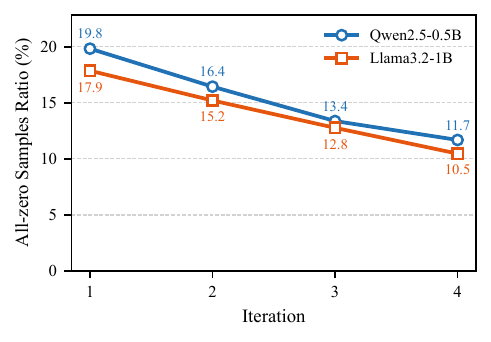}
  \caption{
  All-zero reward sample ratio across training iterations. 
  The ratio indicates that error-guided recycling gradually reduces RL-intractable failures.
  }
  \label{fig:all_zero_ratio_iterations}
\end{figure}

\paragraph{Main Results} We run our iterative SFT-then-RL pipeline for 2 iterations and report the final results in Table~\ref{table:main_results}. 
Our method achieves the best average performance on both Qwen2.5-0.5B and Llama3.2-1B. The gains are consistent across in-domain and out-of-domain benchmarks, suggesting that coordinating SFT and RL through Acquisition, Consolidation, and Recycled Sets improves not only mathematical reasoning but also broader logical reasoning generalization.

\paragraph{Component Ablation.}
We conduct a component ablation study to examine how each stage-specific data strategy contributes to the final performance. 
For space reasons, Figure~\ref{fig:component_ablation} reports the average accuracy across five benchmarks, while the full benchmark-level results are provided in Appendix~\ref{app:component_ablation}. 
We first vary the data recipe in the SFT stage and find that the Medium$\rightarrow$Bridge strategy achieves the strongest average accuracy after initial SFT, demonstrating that our bridge strategy effectively makes hard samples more learnable. Building on this stronger SFT initialization, adding GRPO and the Recycled Set further improves performance. These results suggest that the Acquisition Set, Consolidation Set, and Recycled Set provide complementary benefits in the proposed stage-specific post-training pipeline.

\paragraph{Bridge Operation Ablation}
To examine the contribution of each bridge operation, we conduct an ablation study by removing one operation group at a time. 
We report two metrics: pass@8 on the validation set after the initial SFT stage, and the average accuracy across five benchmarks after two SFT-then-RL iterations. 
As shown in Table~\ref{tab:bridge_operation_ablation}, Full Bridge achieves the best results on both models, indicating that expansion, localization, and compression/drop operations are complementary. 
Removing Expand or Localize weakens the model's ability to learn from jumpy or locally difficult steps, while removing Compress/Drop introduces redundant or low-value reasoning content, especially hurting the final iterative performance on SLMs.

\paragraph{Recycling Effectiveness} Table~\ref{tab:recycling_effectiveness} evaluates how all-zero-reward samples should be handled during RL. Directly applying GRPO after the initial SFT stage yields limited performance, while excluding all-zero-reward samples improves the average accuracy on both models, suggesting that such samples provide sparse and uninformative reward feedback for RL. Moreover, recycling these failures into SFT supervision brings additional gains, confirming that all-zero-reward samples are not useless but should be transformed into learnable supervision rather than directly optimized in RL.

\begin{table}[t]
\centering
\small
\setlength{\tabcolsep}{4.5pt}
\renewcommand{\arraystretch}{1.12}
\begin{tabular}{l l cc}
\toprule
\textbf{Model} & \textbf{Method} & \textbf{pass@8} & \textbf{Avg. Acc.} \\
\midrule
\multirow{5}{*}{\shortstack[l]{Qwen2.5\\0.5B}}
& w/o Expand        & 75.11 & 43.89 \\
& w/o Localize      & 74.86 & 44.35 \\
& w/o Comp./Drop    & 78.35 & 47.67 \\
& Raw Hard          & 75.03 & 44.04 \\
& Full Bridge       & \textbf{79.23} & \textbf{49.69} \\
\midrule
\multirow{5}{*}{\shortstack[l]{Llama3.2\\1B}}
& w/o Expand        & 83.47 & 45.69 \\
& w/o Localize      & 84.29 & 46.22 \\
& w/o Comp./Drop    & 82.33 & 42.07 \\
& Raw Hard          & 82.06 & 40.18 \\
& Full Bridge       & \textbf{85.63} & \textbf{48.53} \\
\bottomrule
\end{tabular}
\caption{
Bridge operation ablation. 
pass@8 is measured on the validation set after initial SFT, and Avg. Acc. is the average accuracy across five benchmarks after 2 SFT-then-RL iterations.
}
\label{tab:bridge_operation_ablation}
\end{table}

\begin{table}[t]
\centering
\small
\setlength{\tabcolsep}{4.5pt}
\renewcommand{\arraystretch}{1.15}
\begin{tabular}{lcc}
\toprule
\textbf{Method} & \textbf{Qwen2.5} & \textbf{Llama3.2} \\
\midrule
Initial SFT + GRPO & 42.16 & 39.21 \\
\quad w/ all-zero exclusion & 44.15 & 41.65 \\
\quad w/ all-zero exclusion + Recycling & \textbf{45.27} & \textbf{43.03} \\
\bottomrule
\end{tabular}
\caption{
Effectiveness of error-guided recycling. 
We report the average accuracy (\%) across five benchmarks. 
Excluding all-zero-reward samples improves GRPO by removing sparse and uninformative reward signals, while recycling these failures into SFT supervision brings further gains.
}
\label{tab:recycling_effectiveness}
\end{table}

\section{Conclusion}
In this paper, we propose a difficulty-aware SFT-then-RL framework for improving SLM reasoning. 
Our method aligns data selection with the distinct roles of each post-training stage: SFT is used to acquire not-yet-mastered reasoning skills, while RL is used to consolidate skills that the model can already partially access. 
To operationalize this idea, we organize training data into stage-specific sets, introduce a Bridge mechanism to transform hard samples into more learnable supervision, and recycle all-zero-reward RL failures into supervised signals through Critique Fine-Tuning. 
Experiments on two SLMs across five reasoning benchmarks demonstrate consistent improvements over representative SFT, distillation, and RL baselines. 
These results suggest that effective SLM post-training requires not only stronger algorithms, but also careful coordination of data difficulty across the SFT and RL stages.

\section*{Limitation}
This work focuses on mathematical and logical reasoning with two representative SLMs, and all training data are derived from GSM8K. 
Although our results show consistent gains across multiple evaluation benchmarks, the effectiveness of the proposed stage-specific data strategy on larger SLMs, non-mathematical domains, and more diverse instruction-following tasks remains to be further explored. 
In addition, both the Bridge module and error-guided recycling rely on teacher-generated supervision and judge-based scoring, which may introduce additional computational cost and inherit biases from the teacher model. 
Future work could investigate more efficient automatic scoring methods and extend the framework to broader reasoning and general-purpose post-training settings.

\bibliography{custom}

\appendix

\section{Motivating Experiment}
\label{app:motivating_experiment}

\subsection{Experimental Details}
\label{app:motivating_experiment_details}

To better understand how sample difficulty affects SFT-stage learning, we conduct a set of controlled motivating experiments on the GSM8K training split. 
We use two representative SLMs, Qwen2.5-0.5B-Instruct and Llama3.2-1B-Instruct, as the base models. 
For each training question, the corresponding base model generates 64 sampled responses, and a rule-based verifier is used to compute the empirical correctness rate. 
We then partition the training samples into three model-specific difficulty groups: samples with correctness rate greater than $0.75$ are labeled as \textit{Simple}, samples with correctness rate lower than $0.25$ are labeled as \textit{Hard}, and the remaining samples are labeled as \textit{Medium}.

\begin{figure}[t]
    \centering
    \includegraphics[width=\columnwidth]{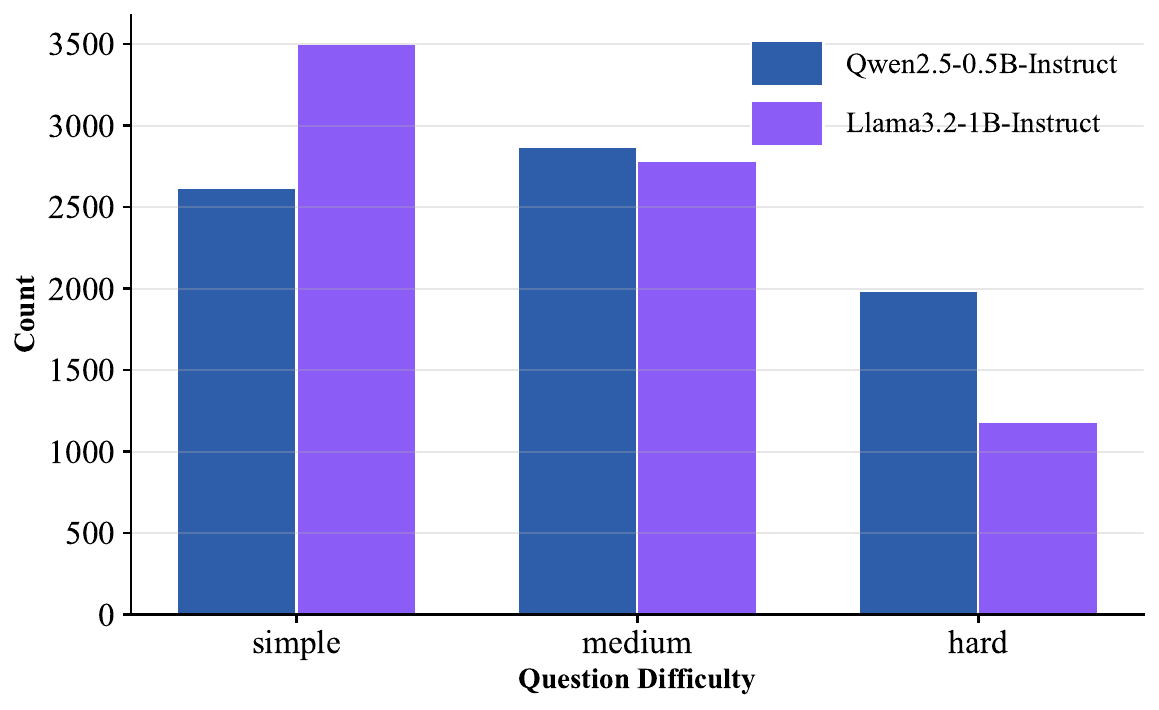}
    \caption{
    Difficulty distribution of GSM8K training samples for each base SLM. 
    The samples are partitioned according to each model's empirical correctness over 64 sampled responses. 
    This figure visualizes the statistics reported in Table~\ref{tab:motivating_difficulty_stats}.
    }
    \label{fig:motivating_difficulty_distribution}
\end{figure}

Since the difficulty partition is model-specific, the same GSM8K question may be assigned to different difficulty groups for different SLMs. 
Table~\ref{tab:motivating_difficulty_stats} and Figure~\ref{fig:motivating_difficulty_distribution} reports the resulting difficulty distribution for the two base models.

\begin{table}[t]
\centering
\small
\setlength{\tabcolsep}{7pt}
\renewcommand{\arraystretch}{1.15}
\begin{tabular}{lcc}
\toprule
\textbf{Difficulty Group} & \textbf{Qwen2.5-0.5B} & \textbf{Llama3.2-1B} \\
\midrule
Simple & 2616 & 3504 \\
Medium & 2869 & 2784 \\
Hard   & 1988 & 1185 \\
\midrule
Total  & 7473 & 7473 \\
\bottomrule
\end{tabular}
\caption{
Model-specific difficulty partition statistics on the GSM8K training split. 
Each sample is assigned to a difficulty group according to the empirical correctness rate of the corresponding base SLM over 64 sampled responses.
}
\label{tab:motivating_difficulty_stats}
\end{table}

After partitioning the training set, we construct five SFT variants for each model: 
\textit{Simple}, which trains only on simple samples; 
\textit{Medium}, which trains only on medium-difficulty samples; 
\textit{Hard}, which trains only on hard samples; 
\textit{Mixed}, which randomly samples from the three difficulty groups according to their original token proportions and shuffles the resulting data to simulate standard SFT; 
and \textit{Medium$\rightarrow$Hard}, which first trains on medium-difficulty samples and then continues training on hard samples as a curriculum-style SFT strategy.

For each variant, we evaluate five supervised assistant-token budgets: 200K, 400K, 600K, 800K, and 1000K tokens. 
All variants are trained under the same base model, optimizer configuration, prompt format, and total assistant-token budget. 
The reasoning trajectory used for SFT is distilled from Qwen-Plus. 
When the total number of available tokens in a difficulty group is insufficient to reach a larger budget, such as 1000K tokens for the hard group, we resample questions from the same difficulty group. 
To avoid simply repeating identical training examples, we re-distill the reasoning trace for the resampled question. 
This makes the repeated sample semantically tied to the same problem while providing a newly generated reasoning trajectory, reducing the risk that the setting degenerates into training on exact duplicate data.

After SFT, we evaluate each trained model using pass@8. 
For each evaluation question, the model samples 64 responses, and pass@8 is computed to measure whether the model can discover at least one correct solution within a small sampling budget. 
This metric is particularly relevant to the SFT-then-RL paradigm, since prior work suggests that pass@$k$ better reflects the exploration potential of a post-SFT model for subsequent RL.

\subsection{Experimental Results}
\label{app:motivating_experiment_results}

\begin{figure*}[t]
  \centering
  \subfloat[Token-budget sensitivity on Qwen2.5-0.5B-Instruct.]{
    \includegraphics[width=0.48\textwidth]{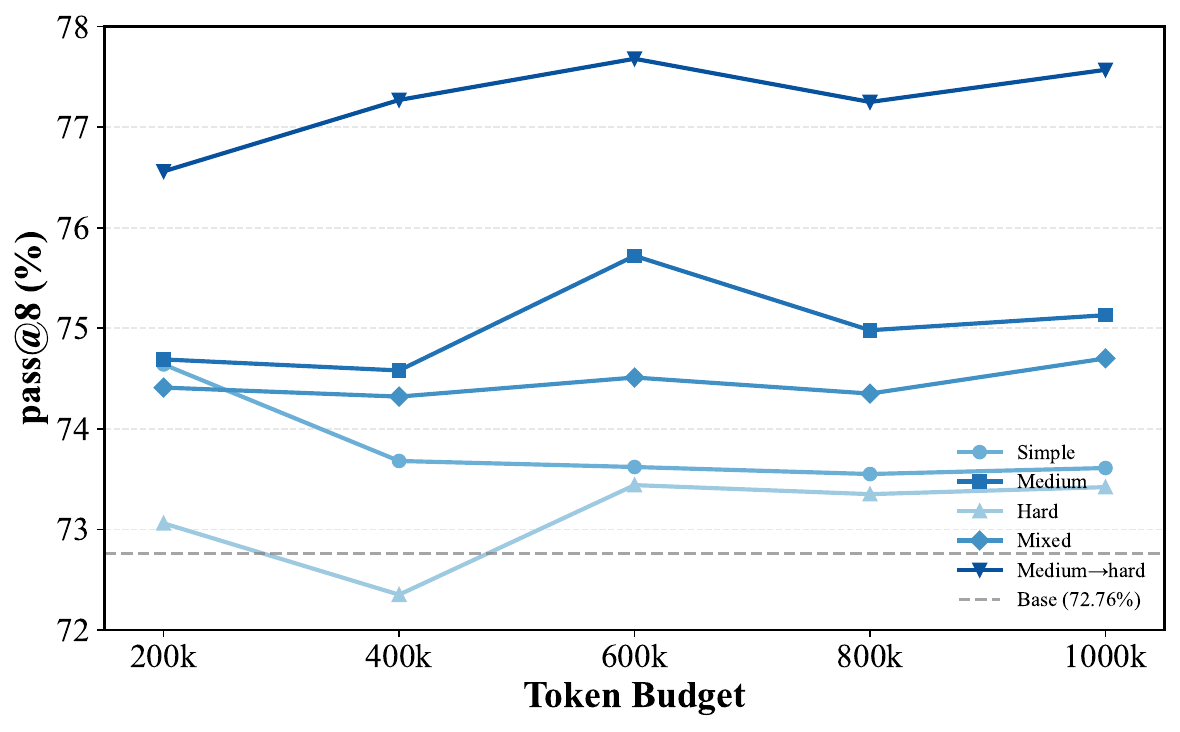}
    \label{fig:qwen_left}
  }
  \hfill
  \subfloat[Average performance across five token budgets.]{
    \includegraphics[width=0.48\textwidth]{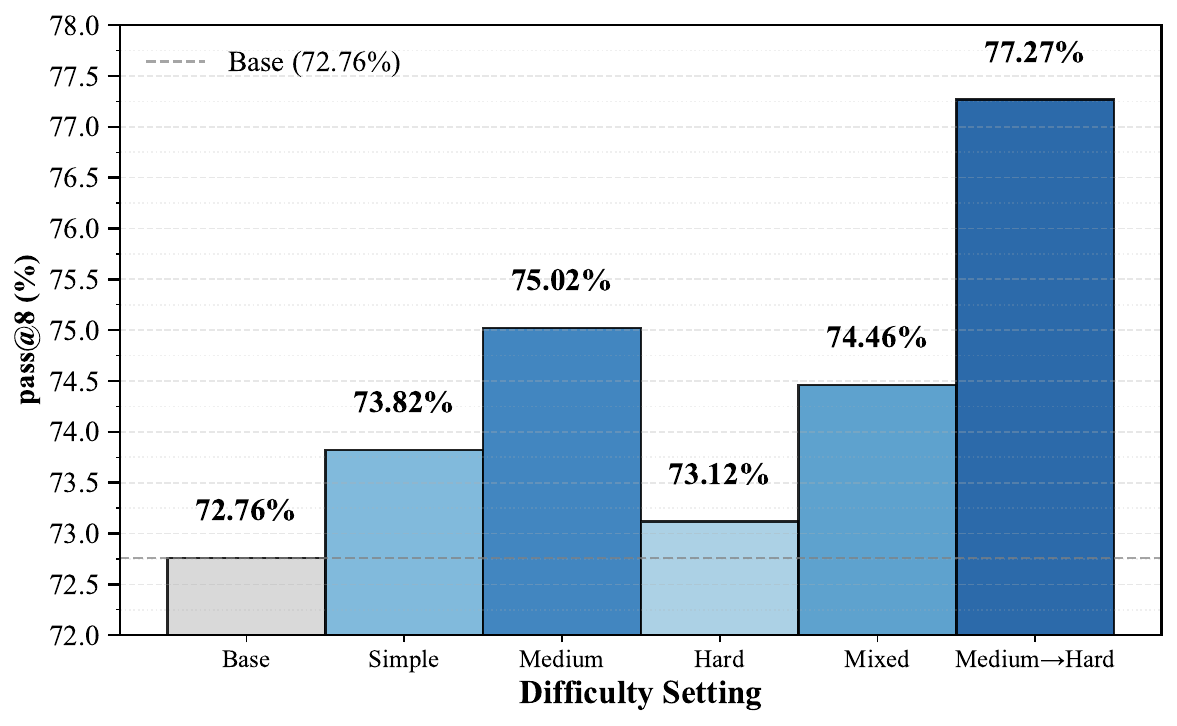}
    \label{fig:qwen_right}
  }
  \caption{
  Motivating difficulty-aware SFT results on Qwen2.5-0.5B-Instruct. 
  The \textit{Medium$\rightarrow$Hard} curriculum achieves the best averaged pass@8, while directly training on hard samples yields the weakest improvement.
  }
  \label{fig:qwen_family}
\end{figure*}

\begin{figure*}[tbp]
  \centering
  \subfloat[Token-budget sensitivity on Llama3.2-1B-Instruct.]{
    \includegraphics[width=0.48\textwidth]{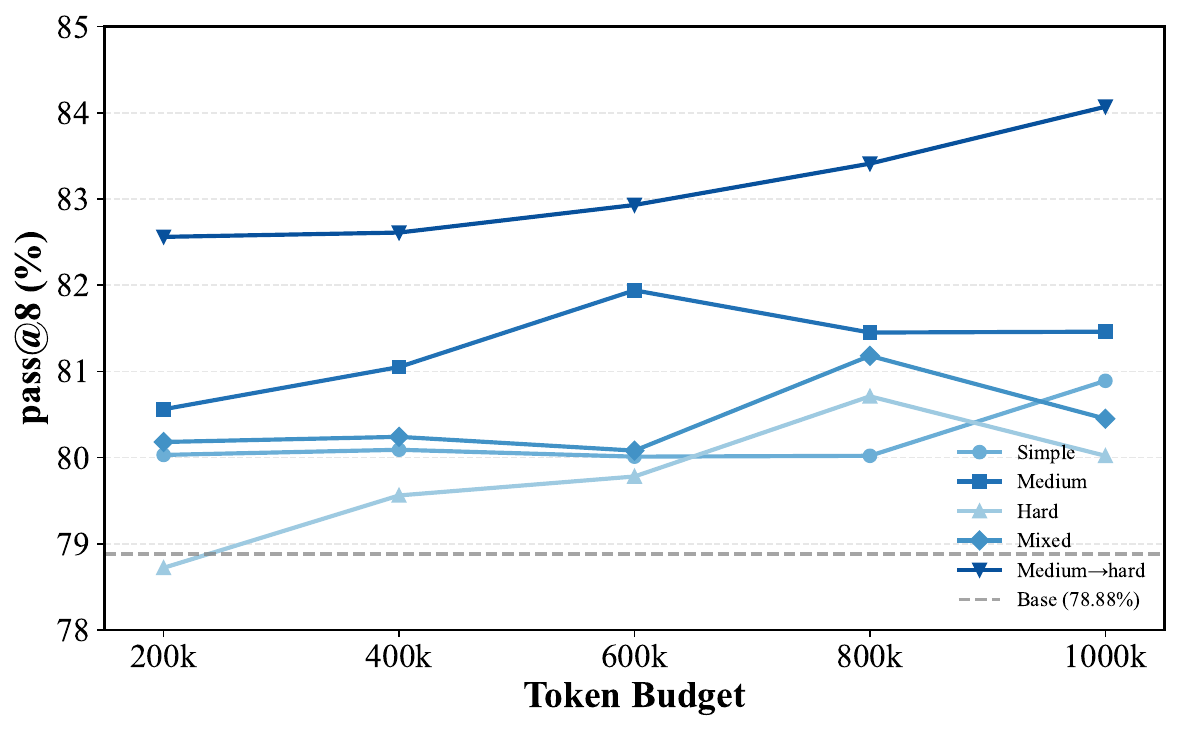}
    \label{fig:llama_left}
  }
  \hfill
  \subfloat[Average performance across five token budgets.]{
    \includegraphics[width=0.48\textwidth]{figures/new_llama_percentage.pdf}
    \label{fig:llama_right}
  }
  \caption{
  Motivating difficulty-aware SFT results on Llama3.2-1B-Instruct. 
  Similar trends are observed: the \textit{Medium$\rightarrow$Hard} curriculum provides the strongest post-SFT pass@8 performance on average.
  }
  \label{fig:llama_family}
\end{figure*}

Figures~\ref{fig:qwen_family} and~\ref{fig:llama_family} report the motivating experiment results on Qwen2.5-0.5B-Instruct and Llama3.2-1B-Instruct, respectively. 
For each model, the left subfigure shows how pass@8 changes under different token budgets, while the right subfigure reports the averaged performance across the five token budgets.

The results show a consistent trend across both model families. 
First, the \textit{Medium$\rightarrow$Hard} curriculum achieves the best average post-SFT performance. 
This suggests that Medium$\rightarrow$Hard samples provide a stable and learnable starting point, while hard samples can further improve the model when introduced after the model has acquired manageable reasoning patterns. 
Therefore, this curriculum provides a stronger initialization for the subsequent SFT-then-RL pipeline.

Second, directly training on \textit{Hard} samples consistently yields the weakest improvement. 
This indicates that although hard samples may contain useful reasoning skills, their raw teacher-distilled reasoning traces are difficult for SLMs to absorb directly. 
This observation motivates our bridge-based transformation, which aims to convert hard samples into a more learnable form before incorporating them into the initial SFT stage.

Third, the \textit{Mixed} setting improves over the base model but is generally weaker than the difficulty-aware curriculum. 
This suggests that simply mixing all difficulty levels according to their original token proportions is less effective than explicitly organizing samples according to their learning function. 
Overall, these motivating experiments support our stage-specific data strategy: SFT benefits from learnable but challenging acquisition data, while hard samples should be incorporated only after being properly scheduled or transformed.

\section{Experimental Settings}
\subsection{Training Hyperparameters}
\label{app:training_hyperparameters}

To facilitate reproducibility, we summarize the key hyperparameters used in the two-stage post-training pipeline. 
Table~\ref{tab:sft_hyperparameters} and Table~\ref{tab:grpo_hyperparameters} reports the most important configurations for both the SFT and GRPO stages. 
For all teacher-generated supervision used in our experiments, including reasoning traces distillation, bridge transformation, and error-guided recycling, we use \texttt{qwen-plus} as the teacher model with temperature $0.7$ and top-$p$ $0.9$.

\begin{table}[t]
\centering
\small
\setlength{\tabcolsep}{4.5pt}
\renewcommand{\arraystretch}{1.15}
\begin{tabular}{ll}
\toprule
\textbf{Category} & \textbf{SFT Setting} \\
\midrule
Framework & VERL SFT trainer \\
Training engine & FSDP \\
Number of GPUs & 2 \\
Training epochs & 1 \\
Train batch size & 8 \\
Micro batch size per GPU & 4 \\
Dynamic batch size & Enabled \\
Padding mode & No padding \\
Truncation strategy & Right truncation \\
Max token length per GPU & 1536 \\
Messages field & \texttt{messages} \\
Learning rate & $2\times10^{-5}$ \\
LR scheduler & Cosine \\
Warmup ratio & 0.2 \\
Minimum LR ratio & 0.1 \\
Weight decay & 0.1 \\
Adam betas & $(0.9, 0.95)$ \\
Gradient clipping & 1.0 \\
Remove padding & Enabled \\
Torch compile & Disabled \\
Save frequency & 100 steps \\
Evaluation frequency & 100 steps \\
Seed & 1111 \\
Resume mode & Disabled \\
\bottomrule
\end{tabular}
\caption{
Key hyperparameters for the supervised fine-tuning stage.
}
\label{tab:sft_hyperparameters}
\end{table}

\begin{table}[t]
\centering
\small
\setlength{\tabcolsep}{4.5pt}
\renewcommand{\arraystretch}{1.15}
\begin{tabular}{ll}
\toprule
\textbf{Category} & \textbf{GRPO Setting} \\
\midrule
Framework & VERL PPO trainer \\
RL algorithm & GRPO \\
Number of GPUs & 4 \\
Training epochs & 1 \\
Train batch size & 16 \\
Prompt length & 1024 \\
Response length & 1024 \\
Actor strategy & FSDP \\
Actor learning rate & $1\times10^{-6}$ \\
PPO mini-batch size & 8 \\
PPO micro-batch size per GPU & 2 \\
Gradient clipping ratio & 0.2 \\
KL loss & Enabled \\
KL coefficient & 0.001 \\
KL loss type & Low-variance KL \\
Rollout engine & vLLM \\
Rollout samples per prompt & 8 \\
Sampling temperature & 0.7 \\
Top-$p$ & 0.9 \\
Rollout dtype & bfloat16 \\
Tensor model parallel size & 4 \\
GPU memory utilization & 0.4 \\
Prefix caching & Disabled \\
Eager execution & Enabled \\
Log-prob micro batch per GPU & 2 \\
Reward filtering & Enabled \\
Filtering metric & All-zero reward \\
Validation before training & Disabled \\
Save frequency & Disabled during training \\
Evaluation frequency & Disabled during training \\
Resume mode & Disabled \\
\bottomrule
\end{tabular}
\caption{
Key hyperparameters for the GRPO stage. 
All-zero-reward filtering is enabled to exclude samples that provide sparse and uninformative reward feedback during policy optimization.
}
\label{tab:grpo_hyperparameters}
\end{table}

\subsection{Dataset Statistics}
\label{app:dataset_statistics}

Table~\ref{tab:dataset_statistics} summarizes the datasets used in our experiments. 
All models are trained exclusively on the GSM8K training split, while evaluation is conducted on both in-domain and out-of-domain benchmarks. 
GSM8K-Platinum is used as the in-domain evaluation set, and MAWPS, SVAMP, and MATH500 are used to evaluate out-of-domain mathematical reasoning. 
We further include LogiQA to assess whether the learned reasoning capability transfers to non-mathematical logical reasoning.

\begin{table}[t]
\centering
\small
\setlength{\tabcolsep}{5pt}
\renewcommand{\arraystretch}{1.15}
\begin{tabular}{l l c}
\toprule
\textbf{Dataset} & \textbf{Description} & \textbf{\# Train / Test} \\
\midrule
GSM8K & Math reasoning & 7473 / 1320 \\
GSM8K-Platinum & Math reasoning & -- / 1209 \\
MAWPS & Math reasoning & -- / 351 \\
SVAMP & Math reasoning & -- / 300 \\
MATH500 & Math reasoning & -- / 500 \\
LogiQA & Logical reasoning & -- / 651 \\
\bottomrule
\end{tabular}
\caption{
Dataset statistics used in our experiments. 
We train only on the GSM8K training split and evaluate on both mathematical and logical reasoning benchmarks.
}
\label{tab:dataset_statistics}
\end{table}

\subsection{Training Objectives}
\label{app:training_objectives}

Our training pipeline uses two objectives: the supervised fine-tuning loss 
$\mathcal{L}_{\mathrm{SFT}}$ for acquisition-oriented learning, and the GRPO loss 
$\mathcal{L}_{\mathrm{GRPO}}$ for reward-based consolidation. 
As shown in Figure~\ref{fig:workflow}, the GRPO objective consists of a PPO-style clipped policy term 
$\mathcal{L}_{\mathrm{CLIP}}$ and a KL regularization term 
$\mathcal{L}_{\mathrm{KL}}$.

\paragraph{Supervised fine-tuning objective.}
Given a supervised dataset $\mathcal{D}_{\mathrm{sup}}$, where 
$\mathcal{D}_{\mathrm{sup}}=\mathcal{D}_{acq}$ for the initial SFT stage and 
$\mathcal{D}_{\mathrm{sup}}=\mathcal{D}_{acq}\cup\mathcal{D}_{rec}^{(t)}$ for later iterations, we optimize the standard next-token prediction loss on assistant tokens:
\begin{equation}
\mathcal{L}_{\mathrm{SFT}}(\theta)
=
-
\mathbb{E}_{(x,y)\sim \mathcal{D}_{\mathrm{sup}}}
\left[
\frac{1}{|y|}
\sum_{k=1}^{|y|}
\log \pi_{\theta}(y_k \mid x,y_{<k})
\right],
\end{equation}
where $x$ denotes the input prompt, $y$ denotes the target assistant response, and $\pi_{\theta}$ denotes the current SLM.

\paragraph{GRPO objective.}
During RL, GRPO is applied to the Consolidation Set $\mathcal{D}_{con}^{(t)}$. 
For each prompt $q$, the old policy $\pi_{\theta_{\mathrm{old}}}$ samples a group of $G$ responses:
\begin{equation}
    \{o_i\}_{i=1}^{G}
    \sim
    \pi_{\theta_{\mathrm{old}}}(\cdot \mid q).
\end{equation}
Each response receives a rule-based reward $r_i$. 
Following GRPO, we compute the group-relative advantage:
\begin{equation}
\hat{A}_i
=
\frac{r_i-\mathrm{mean}(\mathbf{r})}
{\mathrm{std}(\mathbf{r})+\epsilon},
\qquad
\mathbf{r}=[r_1,\ldots,r_G],
\end{equation}
where $\epsilon$ is a small constant for numerical stability.

For the $t$-th token in response $o_i$, we define the policy ratio:
\begin{equation}
\rho_{i,t}(\theta)
=
\frac{
\pi_{\theta}(o_{i,t}\mid q,o_{i,<t})
}{
\pi_{\theta_{\mathrm{old}}}(o_{i,t}\mid q,o_{i,<t})
}.
\end{equation}

The PPO-style clipped term is:
\begin{equation}
\begin{aligned}
\mathcal{L}_{\mathrm{CLIP}}^{(i,t)}(\theta)
=
\min \Big(
&\rho_{i,t}(\theta)\hat{A}_i, \\
&\mathrm{clip}(\rho_{i,t}(\theta),1-\epsilon_c,1+\epsilon_c)\hat{A}_i
\Big),
\end{aligned}
\end{equation}
where $\epsilon_c$ is the clipping coefficient.

The KL regularization term is:
\begin{equation}
\mathcal{L}_{\mathrm{KL}}(\theta)
=
D_{\mathrm{KL}}
\left[
\pi_{\theta}
\|
\pi_{\mathrm{ref}}
\right].
\end{equation}

Then the GRPO maximization objective is:
\begin{equation}
\begin{aligned}
J_{\mathrm{GRPO}}(\theta)
=
\mathbb{E}
\Bigg[
\frac{1}{G}
\sum_{i=1}^{G}
\frac{1}{|o_i|}
\sum_{t=1}^{|o_i|}
\mathcal{L}_{\mathrm{CLIP}}^{(i,t)}(\theta)
-
\beta
\mathcal{L}_{\mathrm{KL}}(\theta)
\Bigg],
\end{aligned}
\end{equation}
where the expectation is over 
$q\sim \mathcal{D}_{con}^{(t)}$ and 
$\{o_i\}_{i=1}^{G}\sim\pi_{\theta_{\mathrm{old}}}(\cdot\mid q)$.

In implementation, we minimize the corresponding loss:
\begin{equation}
\mathcal{L}_{\mathrm{GRPO}}(\theta)
=
-
J_{\mathrm{GRPO}}(\theta).
\end{equation}

\section{Algorithm Details}

\subsection{Bridge-Based Construction}
\label{bridged_samples}
Algorithm~\ref{alg:bridge_construction} summarizes the construction of the bridge-adjusted hard dataset. 
The thresholds $\tau_I$, $\tau_J$, and $\tau_D$ control whether a step is regarded as important, jumpy, or locally difficult, respectively. 
In our experiments, we set the importance and jumpiness thresholds to $\tau_I=0.5$ and $\tau_J=0.5$. 
For the difficulty threshold $\tau_D$, we compute the token-level loss statistics of teacher-generated reasoning trajectory steps on hard samples for each base SLM, and use the model-specific mean loss as the threshold. 
As shown in Table~\ref{tab:difficulty_threshold_stats}, we set $\tau_D=0.99$ for Qwen2.5-0.5B-Instruct and $\tau_D=1.26$ for Llama3.2-1B-Instruct.

Low-importance but non-trivial steps are compressed or dropped, while important steps are kept, expanded, or localized depending on their jumpiness and difficulty. 
In implementation, localize operations may additionally produce local SFT samples. 
For notation simplicity, we include both bridge-adjusted full samples and localized samples in $\mathcal{D}_{bridge}$.

\begin{table}[t]
\centering
\small
\setlength{\tabcolsep}{4.5pt}
\renewcommand{\arraystretch}{1.12}
\begin{tabular}{l c c c c c c}
\toprule
\textbf{Base SLM} & \textbf{Mean} & \textbf{Min} & \textbf{25\%} & \textbf{50\%} & \textbf{75\%} & \textbf{Max} \\
\midrule
Qwen2.5-0.5B & 0.99 & 0.00 & 0.00 & 0.08 & 1.08 & 22.25 \\
Llama3.2-1B & 1.26 & 0.00 & 0.01 & 0.20 & 1.59 & 27.62 \\
\bottomrule
\end{tabular}
\caption{
Token-level loss statistics for teacher-generated reasoning trajectory on hard samples. 
We use the model-specific mean loss as the difficulty threshold $\tau_D$ in the bridge decision rule.
}
\label{tab:difficulty_threshold_stats}
\end{table}

\begin{algorithm}[htbp]
\caption{Bridge-Based Construction of $\mathcal{D}_{bridge}$}
\label{alg:bridge_construction}
\begin{algorithmic}[1]
\Require Hard set $\mathcal{D}_{hard}$, teacher model $T$, current SLM $\pi_{\theta}$, thresholds $\tau_I,\tau_J,\tau_D$
\Ensure Bridge-adjusted hard set $\mathcal{D}_{bridge}$

\State Initialize $\mathcal{D}_{bridge} \leftarrow \emptyset$

\For{each $(q,a^*) \in \mathcal{D}_{hard}$}
    \State Generate teacher reasoning trace $C=[s_1,\ldots,s_m]$ using $T$
    \State Initialize processed trace $\tilde{C} \leftarrow [\ ]$
    \State Initialize local sample set $\mathcal{L} \leftarrow \emptyset$

    \For{$i=1$ to $m$}
        \State $P_i \leftarrow [\tilde{s}_1,\ldots,\tilde{s}_{|\tilde{C}|}]$
        \State Compute step scores $(I_i,J_i,D_i)$ as defined in Section~\ref{sec:bridge_sft}

        \If{$I_i \leq \tau_I$}
            \If{$J_i > \tau_J$ \textbf{or} $D_i > \tau_D$}
                \State $\tilde{s}_i \leftarrow \varnothing$ \Comment{Drop low-value hard/jumpy step}
            \Else
                \State $\tilde{s}_i \leftarrow \mathrm{Compress}(s_i)$
            \EndIf
        \Else
            \If{$J_i \leq \tau_J$ \textbf{and} $D_i \leq \tau_D$}
                \State $\tilde{s}_i \leftarrow s_i$ \Comment{Keep}
            \ElsIf{$J_i > \tau_J$}
                \State $\tilde{s}_i \leftarrow \mathrm{Expand}(q,P_i,s_i)$
                \If{$D_i > \tau_D$}
                    \State Add local sample $(q,P_i,\tilde{s}_i)$ to $\mathcal{L}$
                \EndIf
            \Else
                \State $\tilde{s}_i \leftarrow \mathrm{Localize}(q,P_i,s_i)$
                \State Add local sample $(q,P_i,\tilde{s}_i)$ to $\mathcal{L}$
            \EndIf
        \EndIf

        \If{$\tilde{s}_i \neq \varnothing$}
            \State Append $\tilde{s}_i$ to $\tilde{C}$
        \EndIf
    \EndFor

    \State Add full bridge-adjusted sample $(q,\tilde{C},a^*)$ to $\mathcal{D}_{bridge}$
    \State Add all local SFT samples in $\mathcal{L}$ to $\mathcal{D}_{bridge}$
\EndFor

\State \Return $\mathcal{D}_{bridge}$
\end{algorithmic}
\end{algorithm}

\subsection{Representative Negative Response Scoring}
\label{app:negative_selection}

For each failed sample $(q,a^*) \in \mathcal{D}_{fail}$, all sampled responses are incorrect under the rule-based verifier. 
However, not all incorrect responses are equally useful for teacher-guided error diagnosis. 
Extremely short, degenerate, or unstructured responses often provide little information about where the reasoning process fails. 
Therefore, we select a representative negative response by favoring \textit{near-miss} responses: incorrect responses that still contain a meaningful reasoning attempt and a concrete final-answer prediction.

Given the incorrect response set
\begin{equation}
    \mathcal{O}^{-}(q)=\{o_i \mid r_i=0\}_{i=1}^{G},
\end{equation}
we define a heuristic scoring function $S_{\mathrm{neg}}(\cdot)$ for each response $o \in \mathcal{O}^{-}(q)$:
\begin{equation}
    S_{\mathrm{neg}}(o)
    =
    \lambda_{\ell} S_{\ell}(o)
    +
    \lambda_{s} S_{s}(o)
    +
    \lambda_{f} S_{f}(o),
\end{equation}
where $\lambda_{\ell},\lambda_{s},\lambda_{f}\geq 0$ are weighting coefficients satisfying
$\lambda_{\ell}+\lambda_{s}+\lambda_{f}=3$. In our experiments, we set
$\lambda_{\ell}=\lambda_{s}=\lambda_{f}=1$, assigning equal weight to all three scoring components.

The first component measures whether the response contains sufficient reasoning content:
\begin{equation}
    S_{\ell}(o)
    =
    \min\left(
    \frac{\mathrm{Tok}(o)}{\tau_{\ell}},
    1
    \right),
\end{equation}
where $\mathrm{Tok}(o)$ denotes the number of tokens in $o$, and $\tau_{\ell}$ is a length threshold. 
This term favors responses that are long enough to expose a non-trivial reasoning process, while preventing overly long responses from dominating the score.

The second component measures whether the response has visible reasoning-step structure:
\begin{equation}
    S_{s}(o)
    =
    \min\left(
    \frac{N_{\mathrm{step}}(o)}{\tau_{s}},
    1
    \right),
\end{equation}
where $N_{\mathrm{step}}(o)$ denotes the number of identifiable reasoning units in $o$, and $\tau_s$ is a step-structure threshold. 
This term favors responses whose reasoning process can be decomposed into interpretable steps, making it easier for the teacher model to locate the first reasoning error.

The third component checks whether the response contains a parseable final-answer format:
\begin{equation}
    S_{f}(o)
    =
    \mathbb{I}[\mathrm{Ans}(o)\neq \varnothing],
\end{equation}
where $\mathrm{Ans}(\cdot)$ is the same answer-extraction function used by the rule-based verifier. 
Although the extracted answer is incorrect for all responses in $\mathcal{O}^{-}(q)$, a parseable final answer indicates that the model has completed a full solution attempt rather than producing an incomplete or degenerate output.

The representative negative response is then selected as:
\begin{equation}
    o^{-}
    =
    \arg\max_{o\in \mathcal{O}^{-}(q)}
    S_{\mathrm{neg}}(o).
\end{equation}

This scoring function is designed to select an incorrect but informative response.
By preferring responses with sufficient length, explicit reasoning structure, and a final-answer prediction, the selected $o^{-}$ is more likely to reveal a concrete reasoning gap that can be diagnosed by the teacher model and converted into error-guided SFT supervision.

In implementation, $N_{\mathrm{step}}(o)$ is estimated by counting identifiable reasoning units or step-like cues in the response.

To calibrate the lightweight near-miss scoring function, we compute descriptive statistics of teacher-generated CoT traces for hard samples under each base SLM. 
Table~\ref{tab:teacher_trace_statistics} reports the step-count and token-count distributions. 
Since the traces contain a small number of extremely long cases, we use capped normalization rather than raw length or raw step count. 
Specifically, we set the length threshold $\tau_{\ell}$ to the mean token count of the corresponding base SLM, and set the step threshold $\tau_s$ to the mean step count of the corresponding base SLM. 
Thus, for Qwen2.5-0.5B-Instruct, we use $\tau_{\ell}=242.98$ and $\tau_s=7.36$; for Llama3.2-1B-Instruct, we use $\tau_{\ell}=287.94$ and $\tau_s=13.67$.

\begin{table*}[t]
\centering
\small
\setlength{\tabcolsep}{7pt}
\renewcommand{\arraystretch}{1.15}
\begin{tabular}{l l c c c c c c}
\toprule
\textbf{Base SLM} & \textbf{Metric} & \textbf{Mean} & \textbf{Min} & \textbf{25\%} & \textbf{50\%} & \textbf{75\%} & \textbf{Max} \\
\midrule
\multirow{2}{*}{Qwen2.5-0.5B-Instruct}
& Step count  & 7.36   & 4  & 7   & 7   & 7   & 153  \\
& Token count & 242.98 & 99 & 189 & 214 & 247 & 6763 \\
\midrule
\multirow{2}{*}{Llama3.2-1B-Instruct}
& Step count  & 13.67  & 4  & 7   & 7   & 13  & 162  \\
& Token count & 287.94 & 99 & 183 & 213 & 267 & 5669 \\
\bottomrule
\end{tabular}
\caption{
Statistics of teacher-generated CoT traces for hard samples under each base SLM. 
Step count denotes the number of decomposed reasoning steps, and token count denotes the number of tokens in the teacher-generated reasoning trace. 
We use the model-specific mean token count as $\tau_{\ell}$ and the model-specific mean step count as $\tau_s$ in the near-miss scoring function.
}
\label{tab:teacher_trace_statistics}
\end{table*}

\section{Prompt Templates}
\label{app:prompt_templates}

\subsection{Bridge Trace Construction}
\label{bridge_trace}

This appendix presents the prompt templates used to obtain step-level bridge signals.
For each reasoning step, we query a judge model to estimate two complementary
properties: \textit{importance} and \textit{jumpiness}. The judge is required to
output a discrete score from \scoreset{} only, without any explanation.

\subsubsection{Importance Scoring Prompt}
\label{app:importance_prompt}

The importance score measures how much the deletion of a target reasoning step
damages the remaining chain's ability to naturally support the correct final
answer. Importantly, the judge is instructed not to repair the missing reasoning
using its own external inference.

\begin{promptbox}{System Prompt for Importance Scoring}
You are an objective mathematical reasoning judge.

Your only job is to score the \textbf{IMPORTANCE} of one reasoning step in a correct chain-of-thought.

The score must be based on \textbf{one criterion only}:

\medskip

\emph{How much does deleting this step damage the remaining reasoning chain's ability
to naturally support the given correct final answer?}

\medskip

\textbf{Important rules:}
\begin{enumerate}[leftmargin=1.5em, itemsep=2pt, topsep=2pt]
    \item You must use the provided correct answer as the final target.
    \item You must judge based on the remaining reasoning chain itself.
    \item Do not use your own external reasoning to repair missing logic.
    \item Do not reward a step just because it looks mathematically sophisticated.
    \item Score higher only if removing this step makes the remaining chain much less able
    to naturally justify the correct answer.
\end{enumerate}

Use only one of these five scores:

\medskip

\centerline{\scoreset{}}

\medskip

\textbf{Score interpretation:}
\begin{itemize}[leftmargin=1.5em, itemsep=2pt, topsep=2pt]
    \item \texttt{0.0}: deleting this step has almost no impact.
    \item \texttt{0.25}: slight impact.
    \item \texttt{0.5}: moderate impact.
    \item \texttt{0.75}: large impact.
    \item \texttt{1.0}: critical step; deleting it badly breaks the remaining reasoning.
\end{itemize}

Output only the score number. Do not output any explanation or extra text.
\end{promptbox}

\begin{promptbox}{User Prompt for Importance Scoring}
\textbf{Math Question:}

\placeholder{question}

\medskip

\textbf{Correct Final Answer:}

\placeholder{gold\_answer}

\medskip

\textbf{Full Correct Reasoning Chain (CoT):}

\placeholder{full\_cot}

\medskip

\textbf{Reasoning Chain after Deleting the Target Step:}

\placeholder{cot\_without\_step}

\medskip

\textbf{Deleted Step:}

\placeholder{step\_to\_score}

\medskip

\textbf{Task:}

Judge how much deleting the above step damages the remaining reasoning chain's ability
to naturally support the correct final answer.

\medskip

\textbf{Reminder:}
\begin{itemize}[leftmargin=1.5em, itemsep=2pt, topsep=2pt]
    \item Use the correct final answer as the reference target.
    \item Judge only based on the remaining chain.
    \item Do not fill in missing logic using your own reasoning.
\end{itemize}

Output only one score from \scoreset{}.
\end{promptbox}

\subsubsection{Jumpiness Scoring Prompt}
\label{app:jumpiness_prompt}

The jumpiness score measures how abrupt the transition is from the prior reasoning
context to the current reasoning step. A higher score indicates that more implicit
bridging information is missing, making the transition harder for a small language
model to follow.

\begin{promptbox}{System Prompt for Jumpiness Scoring}
You are a professional evaluator of logical jumpiness in mathematical reasoning.

Your only task is to score how difficult it is for a \textbf{small language model}
to follow the transition from the existing reasoning context to the current reasoning step.

The score must be based on \textbf{one criterion only}:

\medskip

\emph{How much bridge is missing between the prior reasoning context and the current step?}

\medskip

\textbf{Important rules:}
\begin{enumerate}[leftmargin=1.5em, itemsep=2pt, topsep=2pt]
    \item Focus on whether the current step follows naturally from the prior context.
    \item Do not judge whether the current step is mathematically correct in isolation.
    \item Do not use your own reasoning to fill in missing bridges.
    \item A higher score means the current step is more abrupt and harder for a small model to follow.
\end{enumerate}

Use only one of these five scores:

\medskip

\centerline{\scoreset{}}

\medskip

\textbf{Score interpretation:}
\begin{itemize}[leftmargin=1.5em, itemsep=2pt, topsep=2pt]
    \item \texttt{0.0}: no jump; fully smooth and natural.
    \item \texttt{0.25}: very small jump.
    \item \texttt{0.5}: moderate jump; some bridge is missing.
    \item \texttt{0.75}: large jump; small models would struggle.
    \item \texttt{1.0}: extreme jump; the current step is almost disconnected from prior context.
\end{itemize}

Output only the score number. Do not output any explanation or extra text.
\end{promptbox}

\begin{promptbox}{User Prompt for Jumpiness Scoring}
\textbf{Math Question:}

\placeholder{question}

\medskip

\textbf{Prior Reasoning Context:}

\placeholder{prev\_context}

\medskip

\textbf{Current Reasoning Step:}

\placeholder{curr\_step}

\medskip

\textbf{Task:}

Evaluate how jumpy the current step is with respect to the prior reasoning context.

\medskip

\textbf{Reminder:}
\begin{itemize}[leftmargin=1.5em, itemsep=2pt, topsep=2pt]
    \item Judge only whether the transition is too abrupt for a small model.
    \item Do not repair the missing logic by yourself.
\end{itemize}

Output only one score from \scoreset{}.
\end{promptbox}

\subsection{Teacher Prompt for Error-Guided Recycling}
\label{app:teacher_error_prompt}

To construct error-guided recycling samples, we use a teacher model to diagnose
incorrect student responses. Given a question, the ground-truth answer, an
incorrect response, and optionally a reference reasoning chain, the teacher model
is asked to identify the first error and provide a concise corrective signal.
The output is constrained to be a single valid JSON object, which enables
automatic parsing and downstream sample construction.

\begin{promptbox}{System Prompt for Error-Guided Recycling}
You are an experienced math/reasoning teacher who specializes in diagnosing
students' solution errors with surgical precision.

\medskip

You will be given:
\begin{itemize}[leftmargin=1.5em, itemsep=2pt, topsep=2pt]
    \item A problem statement, denoted as \texttt{question}.
    \item The correct answer, denoted as \texttt{ground\_truth}.
    \item One incorrect student response, denoted as \texttt{student\_response}.
    \item Optionally, a reference reasoning chain, denoted as \texttt{reference\_cot}.
\end{itemize}

Your task is to produce a \textbf{structured error analysis}, obeying all of the
following rules:

\begin{enumerate}[leftmargin=1.5em, itemsep=3pt, topsep=2pt]
    \item Output only a single valid JSON object. Do not output explanations,
    Markdown code fences, or any extra text.

    \item The JSON object must contain exactly the following fields, in any order:
    \begin{itemize}[leftmargin=1.5em, itemsep=2pt, topsep=2pt]
        \item \texttt{"first\_error"}:
        a string containing a verbatim excerpt of the student's first incorrect
        step or sentence. The excerpt must be no longer than 120 characters.

        \item \texttt{"error\_type"}:
        a string containing a short label for the error category, such as
        \texttt{"calculation error"}, \texttt{"unit conversion error"},
        \texttt{"misread the question"}, \texttt{"logical leap"}, or
        \texttt{"missing case"}.

        \item \texttt{"why\_wrong"}:
        a string containing one to two sentences explaining why that step is
        incorrect.

        \item \texttt{"missing\_knowledge"}:
        a string describing the key concept or reasoning pattern that the
        student appears to lack.

        \item \texttt{"minimal\_fix"}:
        a string containing a one-sentence hint telling the student how to
        correct that specific step.

        \item \texttt{"correct\_next\_step"}:
        a string containing the correct next step that should immediately follow
        the error point. Do not expand it into a full solution.

        \item \texttt{"short\_correct\_reasoning"}:
        a string containing a complete but concise correct reasoning chain in
        three to five steps. The final line must be exactly
        \texttt{\#\#\#\# <number>}, where \texttt{<number>} is the numeric final
        answer. The number must contain no units, currency symbols, trailing
        punctuation, or thousands separators, and it must equal the provided
        \texttt{ground\_truth}.
    \end{itemize}

    \item Focus on the first error only. Do not enumerate multiple mistakes.

    \item Be concise. Avoid long chains of thought, and avoid re-introducing
    complex higher-order reasoning that the student is unlikely to learn from
    in one pass.

    \item Write all field values in English.
\end{enumerate}
\end{promptbox}

\begin{promptbox}{User Prompt for Error-Guided Recycling}
\textbf{[question]}

\placeholder{question}

\medskip

\textbf{[ground\_truth]}

\placeholder{ground\_truth}

\medskip

\textbf{[student\_response]}

\placeholder{student\_response}

\medskip

\placeholder{optional\_reference\_block}

\medskip

Output the error-analysis JSON now.
\end{promptbox}

\begin{promptbox}{Optional Reference Reasoning Block}
\textbf{[reference\_cot]}

\placeholder{reference\_cot}
\end{promptbox}

\section{Ablation Study}
\subsection{Component Ablation}
\label{app:component_ablation}

Table~\ref{tab:component_ablation_full} reports the complete component ablation results across five evaluation benchmarks. 
The ablation is designed to isolate the contribution of each part of our stage-specific data strategy, including medium-difficulty acquisition, bridge-based hard sample adaptation, GRPO-based consolidation, and error-guided recycling.

\begin{table*}[htbp]
\centering
\small
\setlength{\tabcolsep}{4.5pt}
\renewcommand{\arraystretch}{1.2}
\begin{tabular}{l c cccc c}
\toprule
\multirow{2}{*}{\textbf{Methods}} 
& \textbf{In-Domain} 
& \multicolumn{4}{c}{\textbf{Out-of-Domain}} 
& \multirow{2}{*}{\textbf{Average}} \\
\cmidrule(lr){2-2} \cmidrule(lr){3-6}
& GSM8K-Platinum & MAWPS & SVAMP & MATH500 & LogiQA & \\
\midrule
\multicolumn{7}{l}{\textit{\textbf{\# Qwen2.5-0.5B-Instruct based}}} \\
\midrule
Base & 45.74 & 54.23 & 54.67 & 27.80 & 14.44 & 39.38 \\
SFT-Mixed & 47.97 & 56.06 & 55.33 & 28.60 & 15.21 & 40.63 \\
SFT-Medium & 48.14 & 56.90 & 56.00 & 29.80 & 16.13 & 41.39 \\
SFT-Medium $\rightarrow$ Raw-Hard & 48.80 & 58.03 & 56.67 & 30.00 & 17.05 & 42.11 \\
SFT-Medium $\rightarrow$ Bridged-Hard & 49.46 & 59.44 & 57.33 & 29.60 & 18.89 & 42.94 \\
SFT-Medium $\rightarrow$ Bridged-Hard + GRPO & 49.71 & 59.72 & 58.00 & 30.60 & 22.73 & 44.15 \\
SFT-Medium $\rightarrow$ Bridged-Hard + GRPO + Recycling & \underline{50.38} & \underline{60.56} & \underline{59.33} & \underline{31.80} & \underline{24.27} & \underline{45.27} \\
Full Model & \textbf{54.51} & \textbf{63.94} & \textbf{64.00} & \textbf{33.60} & \textbf{32.41} & \textbf{49.69} \\
\midrule
\multicolumn{7}{l}{\textit{\textbf{\# Llama3.2-1B-Instruct based}}} \\
\midrule
Base & 30.52 & 5.77 & 20.67 & 22.60 & 1.54 & 16.22 \\
SFT-Mixed & 45.90 & 45.63 & 49.33 & 24.20 & 15.36 & 36.08 \\
SFT-Medium & 46.23 & 48.45 & 48.67 & 24.80 & 16.28 & 36.89 \\
SFT-Medium $\rightarrow$ Raw-Hard & 47.48 & 49.86 & 51.37 & 24.40 & 17.05 & 38.03 \\
SFT-Medium $\rightarrow$ Bridged-Hard & 48.22 & 52.68 & 54.00 & 25.20 & 18.13 & 39.65 \\
SFT-Medium $\rightarrow$ Bridged-Hard + GRPO & 49.21 & 58.59 & 55.67 & \underline{25.60} & 19.20 & 41.65 \\
SFT-Medium $\rightarrow$ Bridged-Hard + GRPO + Recycling & \underline{50.29} & \underline{61.41} & \underline{58.33} & 25.00 & \underline{20.12} & \underline{43.03} \\
Full Model & \textbf{54.18} & \textbf{72.68} & \textbf{65.67} & \textbf{26.60} & \textbf{23.50} & \textbf{48.53} \\
\bottomrule
\end{tabular}
\caption{Component ablation of stage-specific data strategies across two SLMs. We report performance (\%) on one in-domain benchmark and four out-of-domain benchmarks. The \textbf{bold} and \underline{underline} indicate the best and second-best results within each model family, respectively.}
\label{tab:component_ablation_full}
\end{table*}

\subsection{Effect of Iterative Training}

Table~\ref{tab:iteration_ablation_qwen} reports the effect of increasing the number of SFT-then-RL iterations on Qwen2.5-0.5B-Instruct. 
The performance improves from the first to the second iteration, indicating that one round of recycling provides useful additional supervision for the next SFT stage. 
However, further iterations lead to performance degradation, suggesting that repeatedly recycling failures may introduce noise or over-specialization rather than continuous improvement. 
Therefore, we use two iterations as the default setting in our main experiments.

\begin{table*}[htbp]
\centering
\small
\setlength{\tabcolsep}{4.5pt}
\renewcommand{\arraystretch}{1.2}
\begin{tabular}{l c cccc c}
\toprule
\multirow{2}{*}{\textbf{Methods}} 
& \textbf{In-Domain} 
& \multicolumn{4}{c}{\textbf{Out-of-Domain}} 
& \multirow{2}{*}{\textbf{Average}} \\
\cmidrule(lr){2-2} \cmidrule(lr){3-6}
& GSM8K-Platinum & MAWPS & SVAMP & MATH500 & LogiQA & \\
\midrule
\multicolumn{7}{l}{\textit{\textbf{\# Qwen2.5-0.5B-Instruct based}}} \\
\midrule
Base & 45.74 & 54.23 & 54.67 & 27.80 & 14.44 & 39.38 \\
Our Model (iter=1) & \underline{50.38} & \underline{60.56} & \underline{59.33} & \underline{31.80} & \underline{24.27} & \underline{45.27} \\
Our Model (iter=2) & \textbf{54.51} & \textbf{63.94} & \textbf{64.00} & \textbf{33.60} & \textbf{32.41} & \textbf{49.69} \\
Our Model (iter=3) & 44.09 & 58.97 & 58.67 & 26.20 & 25.35 & 42.66 \\
Our Model (iter=4) & 41.85 & 55.56 & 57.33 & 22.40 & 21.97 & 39.82 \\
\bottomrule
\end{tabular}
\caption{
Effect of iterative SFT-then-RL training on Qwen2.5-0.5B-Instruct. 
We report accuracy (\%) on one in-domain benchmark and four out-of-domain benchmarks. 
The \textbf{bold} and \underline{underline} indicate the best and second-best results, respectively. 
Performance peaks at the second iteration, showing that iterative recycling is beneficial at early stages but does not yield monotonic gains with more iterations.
}
\label{tab:iteration_ablation_qwen}
\end{table*}

\begin{figure}[htbp]
    \centering
    \includegraphics[width=\columnwidth]{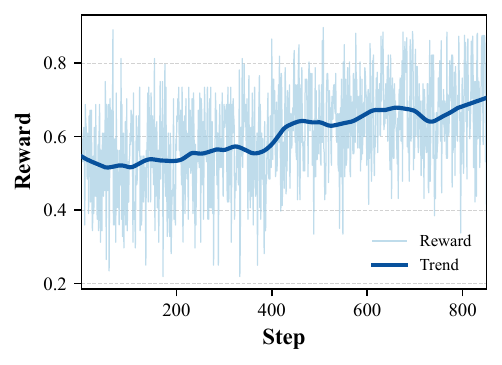}
    \caption{
    Reward dynamics of Llama3.2-1B-Instruct during two iterations of GRPO training. 
    The light curve denotes raw step-level rewards, while the dark curve shows the smoothed trend. 
    Despite high rollout-level variance, the overall reward trend increases throughout training, indicating improved policy optimization under the proposed SFT-then-RL pipeline.
    }
    \label{fig:llama_grpo_reward_curve}
\end{figure}

Figure~\ref{fig:llama_grpo_reward_curve} shows the reward dynamics of Llama3.2-1B-Instruct during 2 iterations of GRPO training. 
Although the step-level rewards exhibit substantial variance due to stochastic rollout sampling, the smoothed trend increases steadily over training. 

\begin{figure}[htbp]
    \centering
    \includegraphics[width=\columnwidth]{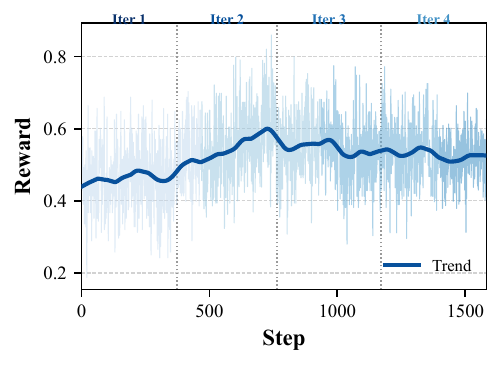}
    \caption{
    GRPO reward curve of Qwen2.5-0.5B-Instruct across 4 training iterations. 
    Vertical dashed lines separate iterations. 
    The smoothed trend rises in the early stages and then decreases but stabilizes.
    }
    \label{fig:qwen_grpo_reward_curve}
\end{figure}

Figure~\ref{fig:qwen_grpo_reward_curve} shows the GRPO reward dynamics of Qwen2.5-0.5B-Instruct across four SFT-then-RL iterations. 
The smoothed reward trend increases in the early iterations, indicating that iterative recycling initially improves the policy's ability to obtain positive rewards. 
However, as the number of iterations continues to increase, the reward trend begins to decline and gradually stabilizes, suggesting that later iterations mainly consolidate the learned behavior rather than bringing continuous reward improvement.

\section{Case Study}
\label{Case Study}
\subsection{Bridge Transformation}
\label{app:bridge_case_study}

Figure~\ref{fig:bridge_case_study} illustrates how the Bridge module transforms a teacher-generated reasoning trace into capacity-aligned supervision. 
For this Llama3.2-1B-Instruct example, we use $\tau_I=0.5$, $\tau_J=0.5$, and $\tau_D=1.26$. 
The difficulty scores $D_i$ are computed as the average token-level loss of each reasoning unit under the current SLM.

\begin{figure}[htbp]
  \centering
  \includegraphics[width=0.95\columnwidth]{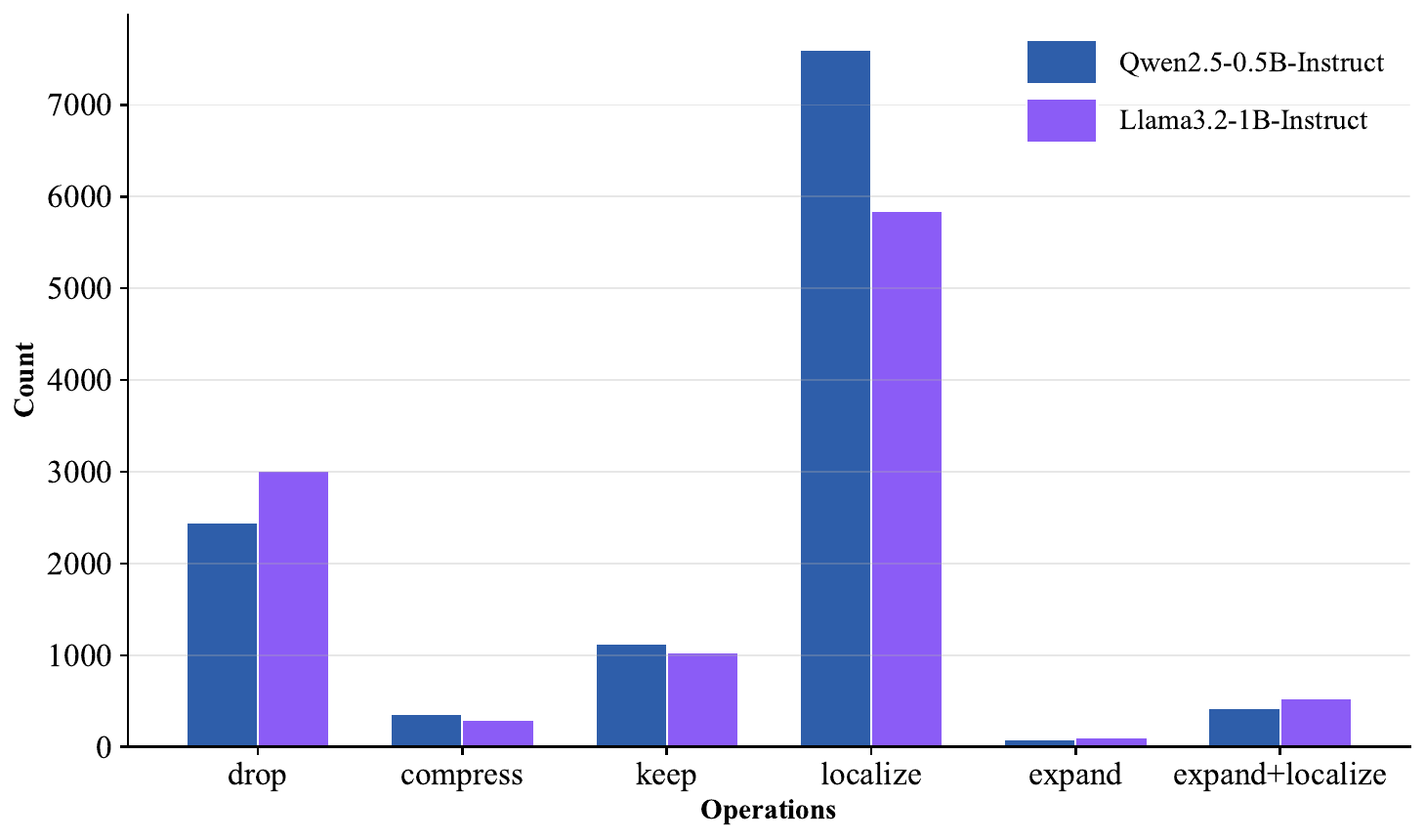}
  \caption{
  Distribution of Bridge actions for Qwen2.5-0.5B-Instruct and Llama3.2-1B-Instruct. 
  The figure reports how often each Bridge operation is applied when transforming hard teacher-generated reasoning traces into capacity-aligned supervision.
  }
  \label{fig:bridge_action_distribution}
\end{figure}

Figure~\ref{fig:bridge_action_distribution} shows the distribution of Bridge actions for Qwen2.5-0.5B-Instruct and Llama3.2-1B-Instruct. 
Across both models, Localize is the most frequently selected operation, indicating that many hard reasoning steps are important but locally difficult for SLMs to imitate directly. 
This suggests that the main challenge in using hard samples is not simply their overall difficulty, but the presence of local reasoning segments that need to be rewritten into more learnable supervision. 
Drop and Keep also account for a noticeable portion of the operations, showing that the Bridge module both removes low-value difficult steps and preserves steps that are already important and learnable. 
In contrast, Compress and Expand are used less frequently, suggesting that redundancy reduction and transition smoothing serve as targeted adjustments rather than dominant transformations.

\begin{table*}[htbp]
\centering
\small
\setlength{\tabcolsep}{4pt}
\renewcommand{\arraystretch}{1.18}
\begin{tabular}{p{0.57\textwidth} c c c p{0.12\textwidth}}
\toprule
\textbf{Original Reasoning Unit} & \textbf{$I_i$} & \textbf{$J_i$} & \textbf{$D_i$} & \textbf{Action} \\
\midrule

\textbf{Unit 1.}
We are told: 
- Carla drank some amount of \textbf{water}, let's call that $w$ ounces. 
- She drank \textbf{three times as much soda minus 6 ounces}, compared to the amount of water she drank. So the amount of \textbf{soda} is:
\[
s = 3w - 6
\]
- Total liquid consumed is 54 ounces:
\[
w + s = 54
\]
& 1.00 & 0.00 & 1.4060 & \textbf{Localize} \\

\midrule

\textbf{Unit 2.}
\#\#\# Step 1: Substitute expression for soda into total equation

Replace $s$ with $3w - 6$:
\[
w + (3w - 6) = 54
\]
& 1.00 & 0.25 & 1.3356 & \textbf{Localize} \\

\midrule

\textbf{Unit 3.}
\#\#\# Step 2: Simplify left-hand side
\[
w + 3w - 6 = 54
\]
\[
4w - 6 = 54
\]
& 0.75 & 0.25 & 0.8654 & \textbf{Keep} \\

\midrule

\textbf{Unit 4.}
\#\#\# Step 3: Solve for $w$

Add 6 to both sides:
\[
4w = 60
\]

Divide both sides by 4:
\[
w = 15
\]
& 1.00 & 0.25 & 1.2773 & \textbf{Localize} \\

\midrule

\textbf{Unit 5.}
\#\#\# Step 4: Check answer

- Water = 15 oz

- Soda = $3(15) - 6 = 45 - 6 = 39$ oz

- Total = $15 + 39 = 54$ oz

$\checkmark$

\texttt{\#\#\#\# 15}
& 0.25 & 0.25 & 0.3166 & \textbf{Compress} \\

\bottomrule
\end{tabular}
\caption{
Original reasoning units and Bridge decisions for the Carla example. 
Each reasoning unit follows the original teacher-generated reasoning trace. 
With thresholds $\tau_I=0.5$, $\tau_J=0.5$, and $\tau_D=1.26$, important but locally difficult units are localized, smooth and learnable units are kept, and low-importance redundant content is compressed.
}
\label{tab:bridge_case_decisions}
\end{table*}

\begin{figure*}[htbp]
\centering
\begin{tcolorbox}[
    enhanced,
    width=0.96\textwidth,
    colback=gray!3,
    colframe=gray!65,
    boxrule=0.6pt,
    arc=2pt,
    left=6pt,
    right=6pt,
    top=6pt,
    bottom=6pt,
    title=\textbf{Bridge Transformation Example},
    fonttitle=\bfseries\small,
    fontupper=\small
]

\textbf{Question}

\begin{tcolorbox}[
    colback=white,
    colframe=gray!35,
    boxrule=0.4pt,
    arc=1pt,
    left=5pt,
    right=5pt,
    top=4pt,
    bottom=4pt
]
Compared to the amount of water she drank, Carla drank three times as much soda minus 6 ounces. 
If she drank 54 ounces of liquid total, how much water did she drink?
\end{tcolorbox}

\vspace{4pt}
\textbf{Bridge-Adjusted Reasoning Trace}

\begin{tcolorbox}[
    colback=blue!3,
    colframe=blue!45,
    boxrule=0.4pt,
    arc=1pt,
    left=5pt,
    right=5pt,
    top=4pt,
    bottom=4pt,
    fontupper=\small
]
We are told: 
- Carla drank some amount of \textbf{water}, let's call that $w$ ounces. 
- She drank \textbf{three times as much soda minus 6 ounces}, compared to the amount of water she drank. So the amount of \textbf{soda} is:
\[
s = 3w - 6
\]
- Total liquid consumed is 54 ounces:
\[
w + s = 54
\]

\#\#\# Step 1: Substitute expression for soda into total equation

Replace $s$ with $3w - 6$:
\[
w + (3w - 6) = 54
\]

\#\#\# Step 2: Simplify left-hand side
\[
w + 3w - 6 = 54
\]
\[
4w - 6 = 54
\]

\#\#\# Step 3: Solve for $w$

Add 6 to both sides:
\[
4w = 60
\]

Divide both sides by 4:
\[
w = 15
\]

\[
\texttt{\#\#\#\# 15}
\]
\end{tcolorbox}

\end{tcolorbox}

\caption{
Bridge transformation case study. 
The original teacher-generated reasoning trace is decomposed into five reasoning units and scored by importance $I_i$, jumpiness $J_i$, and difficulty $D_i$. 
The Bridge module then applies Localize, Keep, or Compress according to Algorithm~\ref{alg:bridge_construction}. 
The final bridge-adjusted trace preserves the core reasoning path while removing redundant verification.
}
\label{fig:bridge_case_study}
\end{figure*}

\begin{figure*}[t]
\centering
\begin{tcolorbox}[
    enhanced,
    width=0.96\textwidth,
    colback=gray!3,
    colframe=gray!65,
    boxrule=0.6pt,
    arc=2pt,
    left=6pt,
    right=6pt,
    top=6pt,
    bottom=6pt,
    title=\textbf{Localized SFT Samples Generated by Bridge},
    fonttitle=\bfseries\small,
    fontupper=\small
]

\textbf{Localize Sample 1: Variable Construction}

\begin{tcolorbox}[
    colback=green!3,
    colframe=green!45!black,
    boxrule=0.4pt,
    arc=1pt,
    left=5pt,
    right=5pt,
    top=4pt,
    bottom=4pt
]
\textbf{Input:} Original question.

\textbf{Target:}
\begin{quote}
\small
We are told: 
- Carla drank some amount of \textbf{water}, let's call that $w$ ounces. 
- She drank \textbf{three times as much soda minus 6 ounces}, compared to the amount of water she drank. So the amount of \textbf{soda} is:
\[
s = 3w - 6
\]
- Total liquid consumed is 54 ounces:
\[
w + s = 54
\]
\end{quote}
\end{tcolorbox}

\vspace{4pt}
\textbf{Localize Sample 2: Substitution}

\begin{tcolorbox}[
    colback=green!3,
    colframe=green!45!black,
    boxrule=0.4pt,
    arc=1pt,
    left=5pt,
    right=5pt,
    top=4pt,
    bottom=4pt
]
\textbf{Input:} Original question + previous reasoning:

\textbf{Target:}
\begin{quote}
\small
\#\#\# Step 1: Substitute expression for soda into total equation

Replace $s$ with $3w - 6$:
\[
w + (3w - 6) = 54
\]
\end{quote}
\end{tcolorbox}

\vspace{4pt}
\textbf{Localize Sample 3: Solving}

\begin{tcolorbox}[
    colback=green!3,
    colframe=green!45!black,
    boxrule=0.4pt,
    arc=1pt,
    left=5pt,
    right=5pt,
    top=4pt,
    bottom=4pt
]
\textbf{Input:} Original question + previous reasoning:

\textbf{Target:}
\begin{quote}
\small
\#\#\# Step 3: Solve for $w$

Add 6 to both sides:
\[
4w = 60
\]

Divide both sides by 4:
\[
w = 15
\]
\end{quote}
\end{tcolorbox}

\end{tcolorbox}

\caption{
Localized SFT samples generated by the Bridge module. 
For important but locally difficult reasoning units, Localize creates targeted supervision that trains the SLM to predict the difficult unit from the question and its preceding context.
}
\label{fig:bridge_localize_samples}
\end{figure*}

\subsection{Error-Guided Recycling}
\label{app:error_guided_case}
Figure~\ref{fig:error_guided_case_nearmiss} and Figure~\ref{fig:error_guided_case_recycled} illustrate how an all-zero-reward failed sample is transformed into three types of recycled SFT supervision. 
The selected near-miss response makes an early variable-definition error, and the teacher-guided diagnosis converts this failure into diagnostic, repair, and new-reasoning trace supervision.

\begin{figure*}[t]
\centering
\begin{tcolorbox}[
    enhanced,
    width=0.96\textwidth,
    colback=gray!3,
    colframe=gray!65,
    boxrule=0.6pt,
    arc=2pt,
    left=6pt,
    right=6pt,
    top=6pt,
    bottom=6pt,
    title=\textbf{Error-Guided Recycling Example: Failed Sample and Near-Miss Response},
    fonttitle=\bfseries\small,
    fontupper=\small
]

\textbf{Original Question}

\begin{tcolorbox}[
    colback=white,
    colframe=gray!35,
    boxrule=0.4pt,
    arc=1pt,
    left=5pt,
    right=5pt,
    top=4pt,
    bottom=4pt
]
In April, Tank gathered 10 more Easter eggs than Emma in their first round of egg hunt. However, Emma gathered twice as many eggs as Tank in their second round of egg hunt, while Tank's total number of eggs in the second round was 20 less than the number she had gathered in the first round. If the total number of eggs in the pile they were collecting with 6 other people was 400 eggs, and Emma gathered 60 eggs in the second egg hunt round, find the number of eggs that the 6 other egg hunters collected?
\end{tcolorbox}

\vspace{4pt}
\textbf{Selected Near-Miss Response}

\begin{tcolorbox}[
    colback=red!3,
    colframe=red!45,
    boxrule=0.4pt,
    arc=1pt,
    left=5pt,
    right=5pt,
    top=4pt,
    bottom=4pt,
    fontupper=\scriptsize
]
We are given a word problem involving egg counts from different rounds and a total pile. 
Let's break it down step by step.

\textbf{Step 1: Define variables.}

Let:
\begin{itemize}
    \item Let $E$ = number of eggs Emma collected.
    \item Tank collected 10 more than Emma in the first round, so Tank $= E + 10$.
    \item Emma collected twice as many as Tank in the second round, so Emma $=2\times(E+10)=2E+20$.
    \item Tank's total in the second round was 20 less than in the first round, so Tank $=E-20$.
    \item Total eggs collected by all 6 people $=400$.
\end{itemize}

\textbf{Step 2: Write an equation for total eggs.}

Total = Emma's first round + Emma's second round + Tank's first round + Tank's second round:
\[
E + (2E+20) + (E-20).
\]

Simplify:
\[
E + 2E + 20 + E - 20 = 4E.
\]
So:
\[
4E = 400,\qquad E = \frac{400}{4}=100.
\]

So Emma collected 100 eggs.

\textbf{Step 3: Find Tank's first-round eggs.}
\[
\text{Tank} = E+10 = 100+10 = 110.
\]

\textbf{Step 4: Find Tank's second-round eggs.}
\[
\text{Emma} = 2E+20 = 200+20 = 220,
\qquad
\text{Tank} = E-20 = 100-20 = 80.
\]

\textbf{Step 5: Total eggs collected by other people.}

Total = Emma's first round + Emma's second round + Tank's first round + Tank's second round:
\[
100 + 220 + 80 + 80 = 400.
\]

\textbf{Final Answer:}
\[
\texttt{\#\#\#\# 400}
\]
\end{tcolorbox}

\end{tcolorbox}

\caption{
Part I of the error-guided recycling case study. 
The all-zero-reward failed sample is reduced to a representative near-miss response \(o^{-}\), which contains a meaningful but incorrect reasoning attempt.
}
\label{fig:error_guided_case_nearmiss}
\end{figure*}

\begin{figure*}[t]
\centering
\begin{tcolorbox}[
    enhanced,
    width=0.96\textwidth,
    colback=gray!3,
    colframe=gray!65,
    boxrule=0.6pt,
    arc=2pt,
    left=6pt,
    right=6pt,
    top=6pt,
    bottom=6pt,
    title=\textbf{Error-Guided Recycling Example: Recycled SFT Supervision},
    fonttitle=\bfseries\small,
    fontupper=\small
]

\textbf{Recycled SFT Samples}

\begin{tcolorbox}[
    colback=blue!3,
    colframe=blue!45,
    boxrule=0.4pt,
    arc=1pt,
    left=5pt,
    right=5pt,
    top=4pt,
    bottom=4pt,
    title=\textbf{1. Diagnostic Supervision: $\mathcal{D}_{diag}^{(t)}$},
    fonttitle=\bfseries\small
]
\textbf{Input:} Question \(q\) + representative negative response \(o^{-}\).

\textbf{Target:}
\begin{quote}
\small
\texttt{\{} \\
\quad \texttt{"error\_type": "misread the question",} \\
\quad \texttt{"first\_error": "Let \$ E \$ = number of eggs Emma collected",} \\
\quad \texttt{"why\_wrong": "The answer incorrectly treats E as Emma's total eggs across both rounds, but the problem gives Emma's second-round count explicitly (60 eggs) and relates rounds separately; E must represent only first-round eggs or be defined per round."} \\
\texttt{\}}
\end{quote}
\end{tcolorbox}

\vspace{4pt}
\begin{tcolorbox}[
    colback=green!3,
    colframe=green!45!black,
    boxrule=0.4pt,
    arc=1pt,
    left=5pt,
    right=5pt,
    top=4pt,
    bottom=4pt,
    title=\textbf{2. Repair Supervision: $\mathcal{D}_{repair}^{(t)}$},
    fonttitle=\bfseries\small
]
\textbf{Input:} Question \(q\) + correct prefix \(p_{\mathrm{pre}}\) + first error \(f_{\mathrm{err}}\) + minimal-fix hint \(h_{\mathrm{fix}}\).

\textbf{First error:}
\[
\text{Let } E = \text{ number of eggs Emma collected.}
\]

\textbf{Minimal-fix hint:}
Define separate variables for each person's eggs per round, or let \(E_1\) denote Emma's first-round eggs, since her second-round eggs are given as 60.

\textbf{Target correct next step:}
\begin{quote}
\small
Let \(E_1\) = Emma's first-round eggs. Then Tank's first-round eggs \(= E_1 + 10\).
\end{quote}
\end{tcolorbox}

\vspace{4pt}
\begin{tcolorbox}[
    colback=yellow!6,
    colframe=orange!65!black,
    boxrule=0.4pt,
    arc=1pt,
    left=5pt,
    right=5pt,
    top=4pt,
    bottom=4pt,
    title=\textbf{3. New-Reasoning Trace Supervision: $\mathcal{D}_{new}^{(t)}$},
    fonttitle=\bfseries\small
]
\textbf{Input:} Original question \(q\).

\textbf{Target new reasoning trace \(c_{\mathrm{new}}\):}
\begin{quote}
\small
Emma gathered 60 eggs in round 2. Since Emma gathered twice as many as Tank in round 2, Tank gathered \(60 \div 2 = 30\) eggs in round 2. 
Tank's round 2 count was 20 less than her round 1 count, so Tank's round 1 count \(= 30 + 20 = 50\). 
Then Emma's round 1 count \(= 50 - 10 = 40\). 
So Emma's total is \(40 + 60 = 100\), and Tank's total is \(50 + 30 = 80\). 
Together they gathered \(100 + 80 = 180\). 
The total pile is 400, so the other 6 hunters gathered \(400 - 180 = 220\).

\texttt{\#\#\#\# 220}
\end{quote}
\end{tcolorbox}

\end{tcolorbox}

\caption{
Part II of the error-guided recycling case study. 
The teacher diagnosis converts the near-miss response into three recycled SFT samples: diagnostic supervision \(\mathcal{D}_{diag}^{(t)}\), repair supervision \(\mathcal{D}_{repair}^{(t)}\), and new-reasoning trace supervision \(\mathcal{D}_{new}^{(t)}\).
}
\label{fig:error_guided_case_recycled}
\end{figure*}

\end{document}